\documentclass{article}

\usepackage[final]{neurips_2019}

\usepackage[utf8]{inputenc} 
\usepackage[T1]{fontenc}    
\usepackage{hyperref}       
\usepackage{url}            
\usepackage{booktabs}       
\usepackage{amsfonts}       
\usepackage{nicefrac}       
\usepackage{microtype}      

\usepackage{times}  
\usepackage{helvet} 
\usepackage{courier}  
\usepackage{graphicx}
\graphicspath{{Plots/}}
\usepackage{amsmath}
\DeclareMathOperator{\E}{\mathbb{E}}
\usepackage{algorithm,algorithmic}

\usepackage{hyperref}

\usepackage{wrapfig}
\usepackage{subcaption}
\usepackage{subcaption} 
\usepackage{amssymb,amsmath}
\usepackage{amsthm}
\usepackage{bm}
\usepackage{dsfont}

\def\expectation{\mathbb{E}}

\title{Off-Policy Policy Gradient Algorithms \\
by Constraining the State Distribution Shift}

\def\sharedaffiliation{%
\end{tabular}\\
\begin{tabular}{c}}
\newcommand*\samethanks[1][\value{footnote}]{\footnotemark[#1]}

\author{
Riashat Islam$^1$\\
\And
Komal K. Teru$^1$\thanks{Equal contribution} \\
\And
Deepak Sharma$^1$\samethanks \\
\And
Joelle Pineau$^{1,2}$\\
\vspace{10pt}
\sharedaffiliation
$^1$Mila, McGill University\\
$^2$Facebook AI Research, Montreal
}

\begin{document}

\maketitle

\begin{abstract}

Off-policy deep reinforcement learning (RL) algorithms are incapable of learning solely from batch offline data without online interactions with the environment, due to the phenomenon known as \textit{extrapolation error}. This is often due to past data available in the replay buffer that may be quite different from the data distribution under the current policy. We argue that most off-policy learning methods fundamentally suffer from a \textit{state distribution shift} due to the mismatch between the state visitation distribution of the data collected by the behaviour and target policies. This data distribution shift between current and past samples can significantly impact performance of most modern off-policy based policy optimization algorithms. In this work, we first do a systematic analysis of state distribution mismatch in off-policy learning, and then develop a novel off-policy policy optimization method to constraint the state distribution shift. To do this, we first estimate the state distribution based on features of the state, using a density estimator and then develop a novel constrained off-policy gradient objective that minimizes the state distribution shift. Our experimental results on continuous control tasks show that minimizing this distribution mismatch can significantly improve performance in most popular practical off-policy policy gradient algorithms.

\end{abstract}

\section{Introduction}

Most off-policy deep RL algorithms that seek to train agents from a fixed batch setting are forced to rely on online sample collection (using a behaviour policy as an instance of the current policy, e.g with noise added to action space) since the data under the behaviour policy can often be uncorrelated with the data that would have been collected under the current policy. Due to this, most deep RL algorithms cannot rely on purely offline batch data. This is fundamentally due to the data distribution shift leading to an extrapolation error as coined by \cite{bcqFujimoto2018}. 

In this work, we propose a constrained policy optimization method based on minimizing the state distribution shift between the target and behaviour policy state distribution. Policy gradient methods such as the Deep Deterministic Policy Gradients (DDPG) \citep{lillicrap2015continuous}, Twin Delayed Deterministic Policy Gradients (TD3) \citep{TD3} and Soft Actor-Critic (SAC) \citep{SAC} are very popular for solving continuous control tasks, and the off-policy nature of these algorithms due to the ability to learn from replay buffer data often makes them more sample efficient compared to the on-policy counterparts such as TRPO \citep{trpo}. However, these methods are incapable to learn solely from off-policy data, and often requires online samples in a growing batch off-policy setting for good performance. We first analyse the sample efficiency of popular off-policy based policy gradient methods such as DDPG, TD3, and SAC \citep{lillicrap2015continuous,TD3,SAC}. We interpolate between pure batch setting and online sample collection to investigate how much these methods can rely on past data. We empirically show that they have to collect data online to perform well, in a growing batch off-policy setting. We hypothesize that this is mostly because of a lack of state distribution correction in off-policy actor-critic methods, resulting in a state distribution shift which further amplifies the effects of extrapolation error. Although recent works have aimed towards state distribution correction \citep{state_correction}, such methods are often not scalable to the deep RL continuous control tasks based on popular value-gradient based algorithms \citep{silver2014deterministic}.

We propose an algorithm that constrains the state distribution to account for the mismatch in the data distribution under the current policy and past behaviour policies. To do this, we first present a novel method to estimate the state distribution with state features based on the state occupancy measure using a density estimator (e.g a variational auto-encoder \citep{VAE}). We then constrain the policy gradient update to minimize the KL divergence between the behaviour and target state distributions. Based on our proposed state distribution constrained policy gradient algorithm, we find that correcting for the mismatch in the data distribution can significantly improve performance in off-policy based methods. The \textbf{key contributions} of our work are as follows :

\begin{itemize}

    \item  We analyse the sample efficiency of popular off-policy actor-critic algorithms. We train agents using sample buffers that interpolate between current and past samples. We also train them in a pure batch setting, in which the replay buffer is fixed. We find that even off-policy based methods rely on a growing batch of online samples. We hypothesize that this is due to the mismatch in state distributions under the behaviour and learned policies.
    
    \item We present a novel method to estimate the state distribution using state features based on the state occupancy measure. We use a parameterized density estimator that directly uses the state features to estimate the state distribution $d_{\pi}(s)$, where we use a policy architecture to output the state features based on state visitation. This coupling ensures that we can find a gradient estimate $\nabla_{\theta} d_{\pi_{\theta}}(s)$ w.r.t to the policy parameters $\theta$ for the state distribution, even though we use a separately parameterized density estimator with parameters $\phi$. 
    
    \item We then present an off-policy policy gradient algorithm that relies on estimating the state distributions under the behaviour and target policies, to constrain and minimize the state distribution mismatch. Our proposed algorithm uses a KL-regularized constraint for the state distributions $KL(d_{\mu} || d_{\pi})$, to minimize the state distribution shift and improve the policy by moving away from the state distribution under the behaviour policy. Our algorithm provides a similar trust region constraint as in \cite{trpo} based on KL between the state instead of policy distributions.
    
\end{itemize}



\section{Preliminaries}

We consider a finite horizon MDP $M = <\mathcal{S}, \mathcal{A}, \mathcal{P}, r, \gamma>$ with a continuous state space $\mathcal{S}$ and continuous action space $\mathcal{A}$, with transition dynamics $\mathcal{P}$, rewards $r$ and discount factor $\gamma$. The goal is to find a parameterized policy $\pi_{\theta}(a,s)$ which can maximize the expected discounted return  $J(\pi) = \E_{s \sim d_{\pi}(s), a \sim \pi(a|s)} [\sum_{t=0}^{\infty} \gamma^{t} r_{t}]$, where $d_{\pi}(s)$ is the stationary state distribution, given by $d_{\pi}(s) = \frac{1}{\sum_{t=0}^{T} \gamma^{t}} \sum_{t=0}^{T} \gamma^{t} d^{t}_{\pi}(s)$. In contrast, the discounted state distribution with discounting applied to future states is given by $d_{\pi}(s) = (1 - \gamma) \sum_{t=0}^{\infty} \gamma^{t} \mathop{P}(s_t=s \mid s_{0})$. We mostly consider policy gradient based methods, based on the policy gradient theorem \citep{sutton2000policy}. The policy gradient theorem is used to maximize the cumulative return objective function given by $\E_{s \sim d_{\pi}(s), a \sim \pi(a,s)} [ \nabla_{\theta} \log \pi_{\theta}(a|s) Q^{\pi}(s,a) ]$. In off-policy based policy gradient methods, a different behaviour policy, denoted by $\mu(a,s)$, is used to collect the data to obtain the policy gradient with an importance sampling based correction for the mismatch in action distributions \citep{precup2000eligibility}. Most off-policy based actor-critic algorithms are derived from the Off-PAC algorithm \citep{degris2012off}, where the policy gradient is given by $\nabla_{\theta}J(\theta) = \E_{s \sim d_{\mu}(s), a \sim \mu(a,s)} [ \frac{\pi(a,s)}{\mu(a,s)} \nabla_{\theta} \log \pi_{\theta}(a|s) Q^{\pi}(s,a) ]$. In practice, an exploratory behaviour policy $\mu(a,s)$ is used which is often coupled with the target policy, given by $\mu(a,s) = \pi(a,s) + \epsilon$ for continuous action space MDPs. 


\section{Distribution Shift in Off-Policy Methods}

\begin{figure*}[!htb]
    \centering
    \includegraphics[width=.31\textwidth]{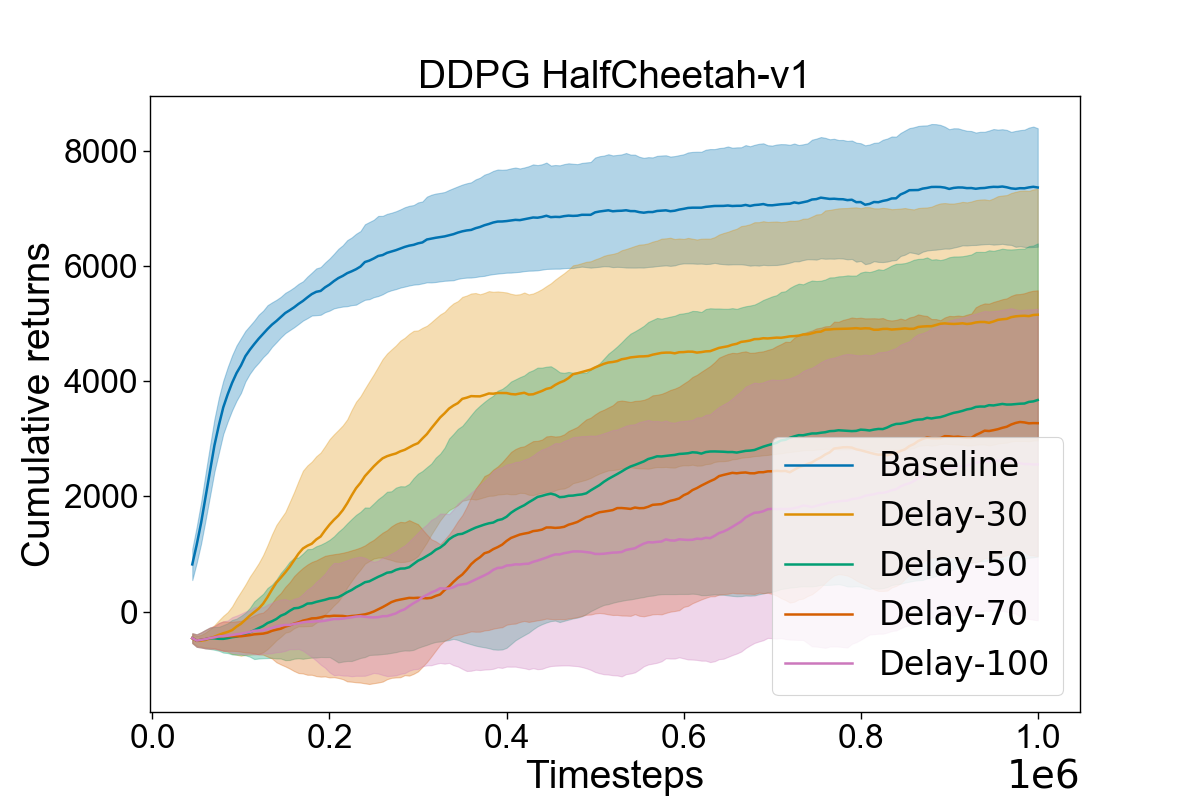}
    \includegraphics[width=.31\textwidth]{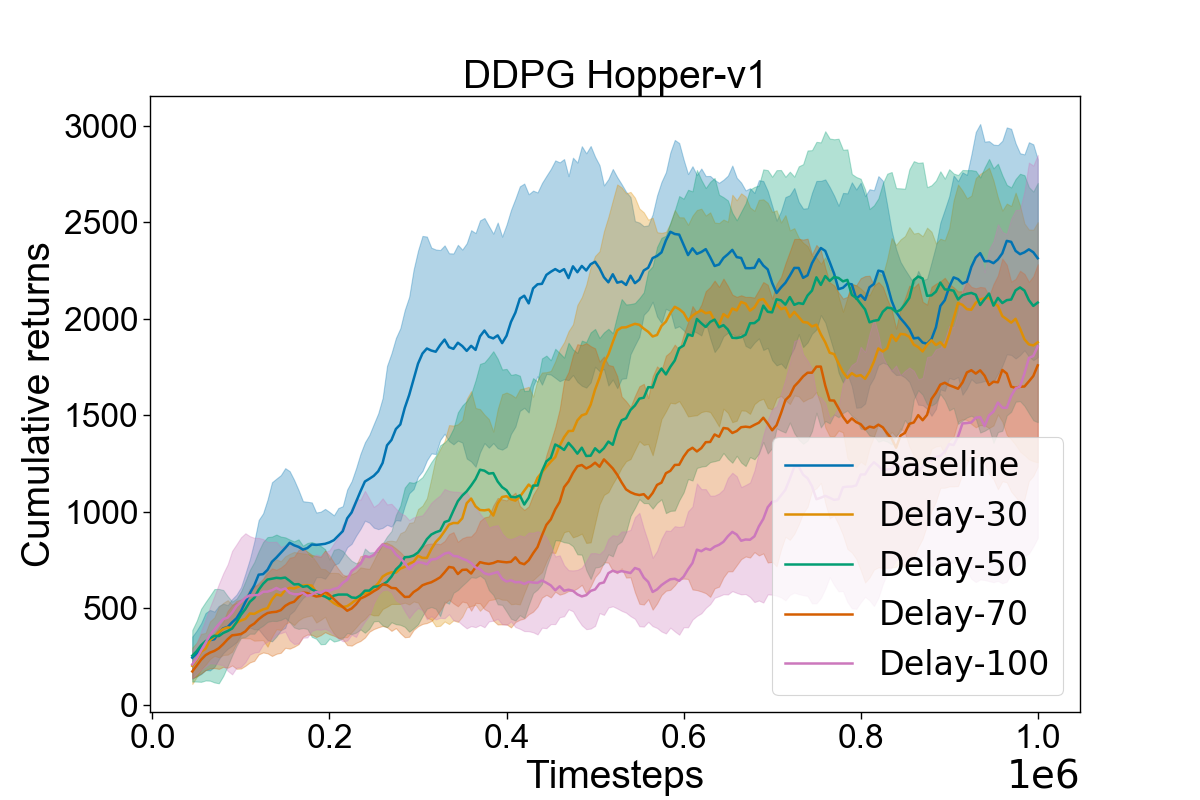}
    \includegraphics[width=.31\textwidth]{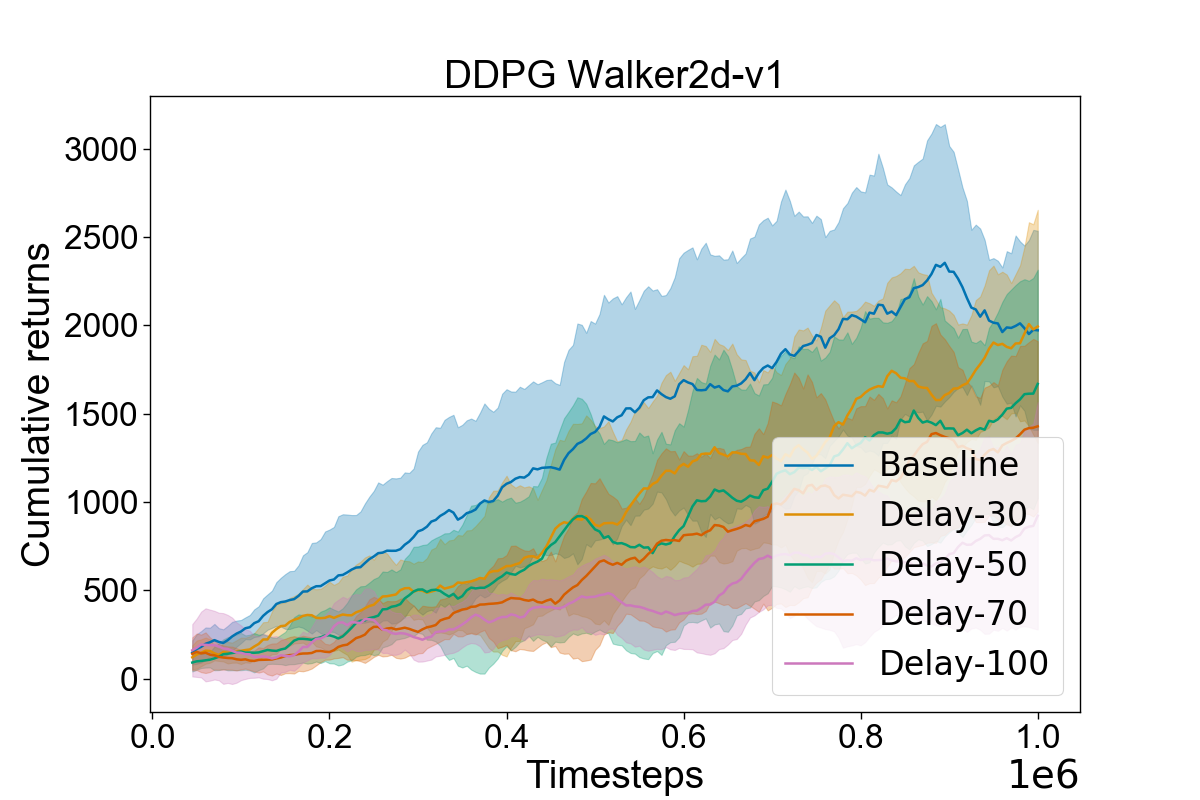}
    \caption{Reward curves for DDPG with delayed buffer sampling average over \textit{10} seeds. The baseline corresponds to $d = 0$ delay. Delay is in episodes. We find that the more delayed samples we use from the replay buffer, the worse the performance gets, and it is always best to use random batch selection containing past and present policies.}
    \label{fig:ddpg_delay}
\end{figure*}

\begin{figure*}[!htb]
    \centering
    \includegraphics[width=.31\textwidth]{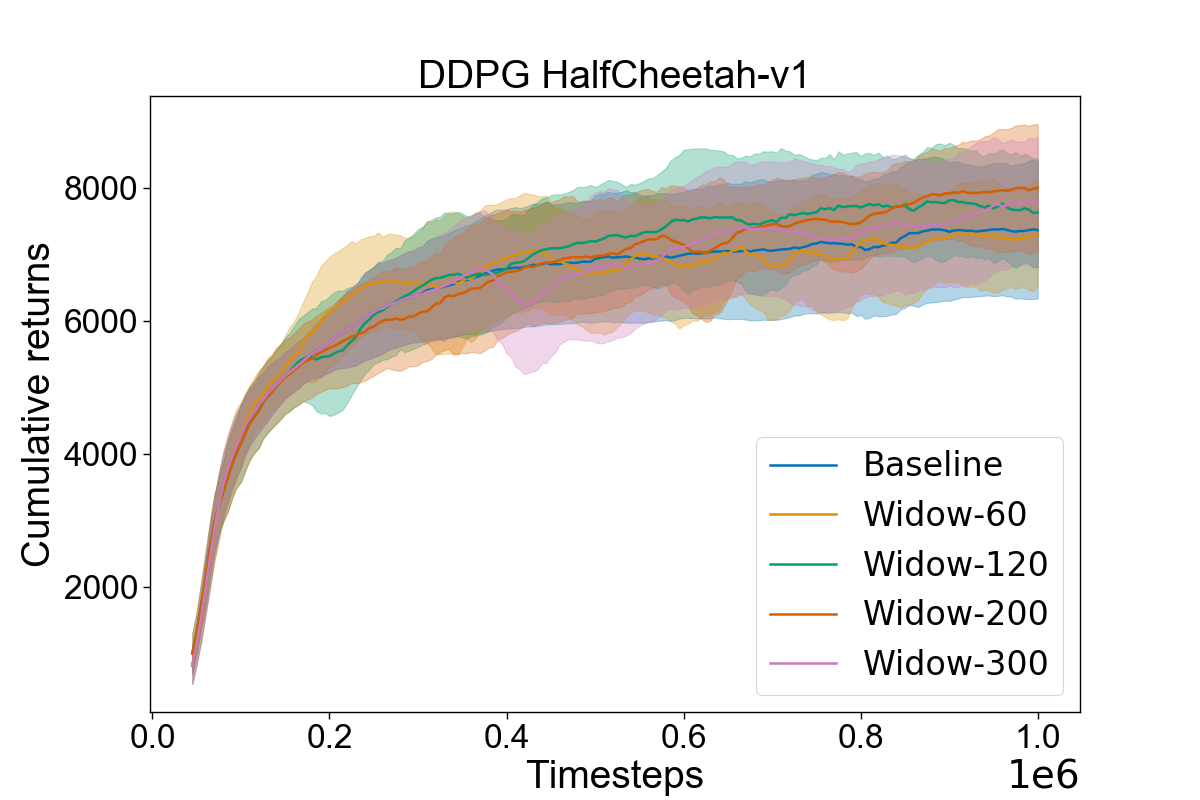}
    \includegraphics[width=.31\textwidth]{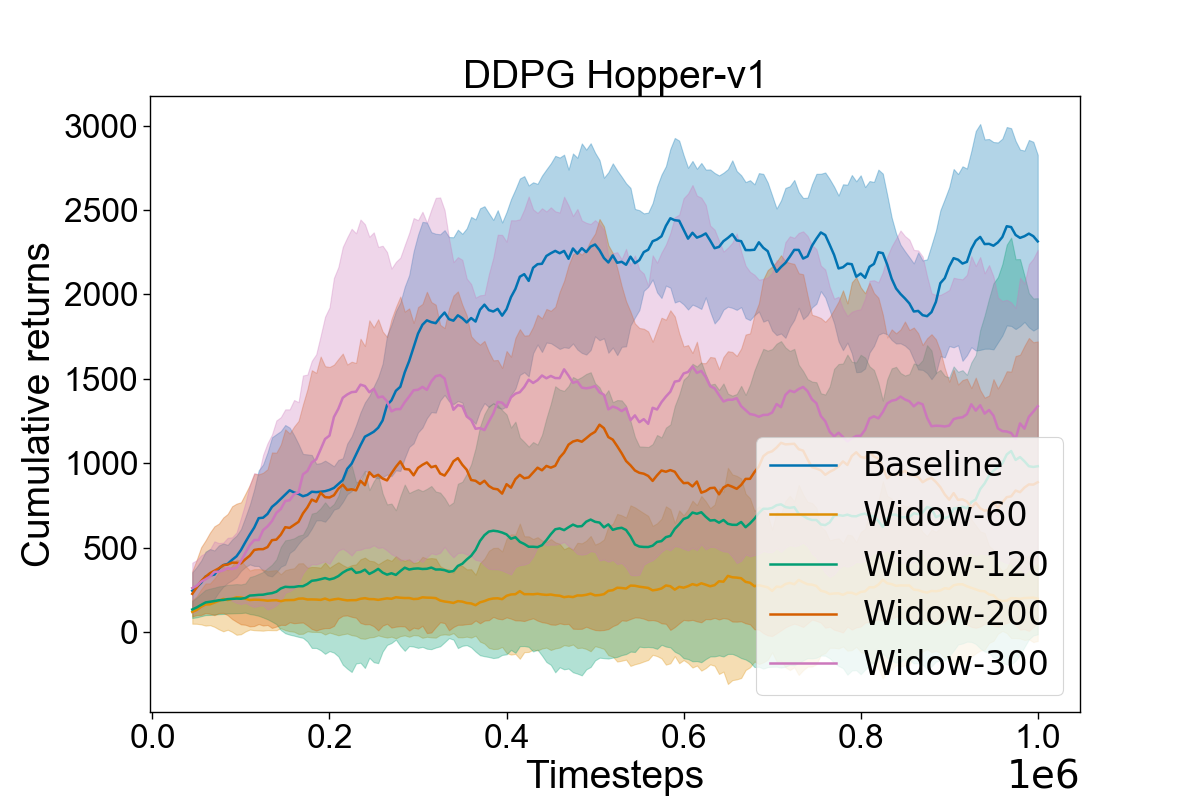}
    \includegraphics[width=.31\textwidth]{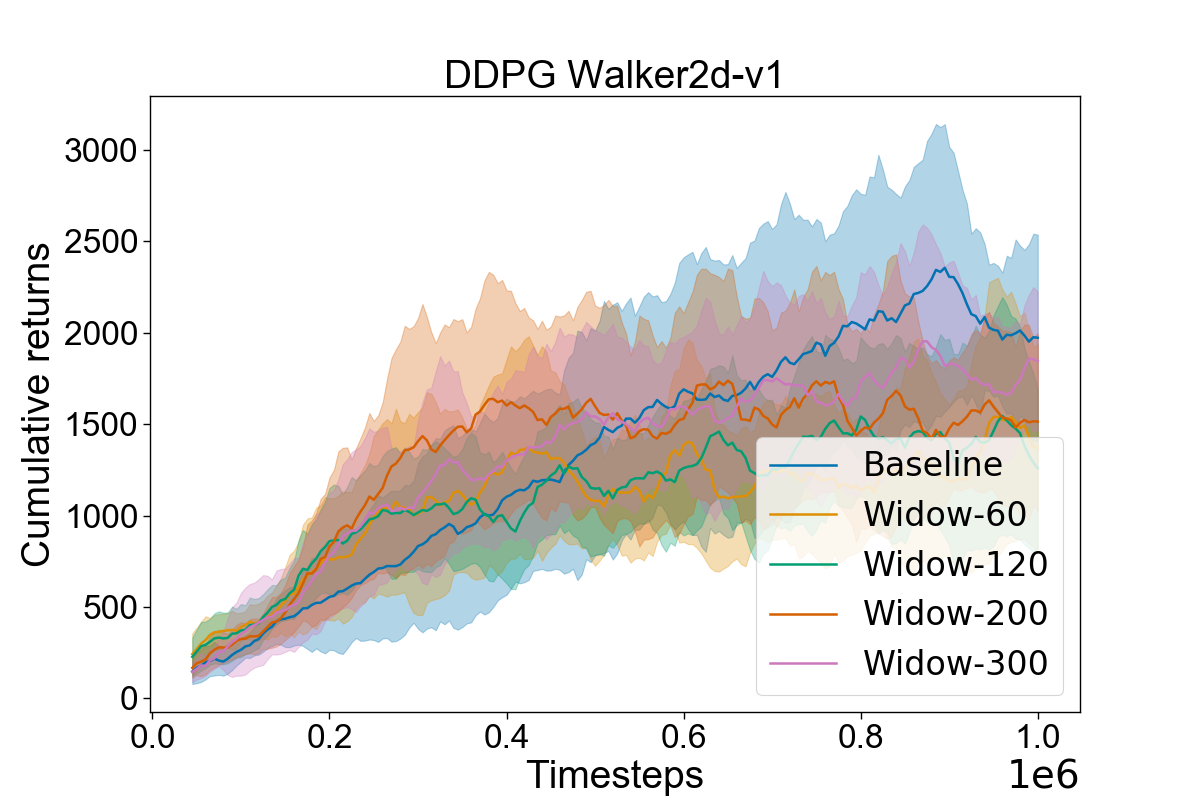}
    \caption{\small{Reward curves for DPPG with windowed buffer sampling averaged over \textit{10} seeds. Windowing is in episodes. The baseline corresponds to window size of $W = 10000$ episodes. We find that using the most recent samples in the replay buffer is often most useful, as these samples have less distribution shift compared to the data distribution under the target policy. This is as expected, since in most of these algorithms we often use a behaviour policy $\mu(a,s) = \pi(a,s) + \epsilon$ that is only separated by a noise $\epsilon$ in the action selection}}
    \label{fig:ddpg_window}
\end{figure*}

We first investigate to what extent state distribution shift matters in off-policy based methods, by interpolating between very old data from the buffer to the most recent data collected under the behaviour policy. We find that the data distribution shift can significantly impact performance in policy updates, where very old samples from the batch data can significantly worsen performance. Our analysis suggests that off-policy based methods are highly dependent on the online most recent samples collected under the behaviour policy, especially since the behaviour policy is often coupled with the target policy and separated by noise in the action space. We systematically analyse the impact of state distribution shift in off-policy gradient methods, by changing the sampling scheme used to collect batches of data from the replay buffer. We motivate the impact of state distribution shift, by introducing two new sampling schemes from the replay buffer. We use popular algorithms namely Deep Deterministic Policy Gradient Algorithm (DDPG) \citep{todorov2012mujoco} and its variants TD3 \citep{bcqFujimoto2018} and Soft Actor-Critic \citep{SAC} and analyse the impact of state distribution shift in continuous control Mujoco benchmark tasks. Additional experiment results are included in the appendix. 

\textit{Sampling from Delayed Replay Buffer} : We sample transitions that were made at least $d$ episodes ago. We use samples of data that were collected with older policies, which may be significantly different compared to the current policy. This lets us study the impact of distribution shift in these methods. \textit{Sampling from Windowed Replay Buffer : } We can also sample transitions only from a window of the latest $W$ episodes or roll-outs. This lets us study the extent to which off-policy methods rely on recent samples. 

\textit{Experiment Results with Delayed Replay Buffer:}  From Figure 1, we observe that the baseline (corresponding to delay = 0) consistently performs the best. As we increase the delay, we are effectively increasing the state distribution shift. When the distribution of data in the buffer does not correspond with the distribution induced by the current policy, this leads to a poor value function approximation. Therefore, our experimental results in Figure 1 verify the hypothesis that increase in distribution shift degrades the performance of the policy since policy improvement with the policy gradient update relies on out of state-distribution samples, which may not be visited by the target policy at all. 

\textit{Experiment Results with Windowed Replay Buffer:} are as shown in Figure 2. We hypothesize that when the data used for state-action value updates is inadequate, the values of missing state-action pairs are extrapolated by the critic, causing errors in the action values. Bootstrapping with erroneous values propagates the error to all state-action values.

\textit{Summary :} From our experimental studies with the different sampling schemes from the replay buffer, we can conclude that the state distribution shift indeed can degrade performance of off-policy policy gradient algorithms. When using \textit{delayed buffer} for sampling batches of data, we find that the baseline with random sampling performs best since it can explicitly use both the online and offline samples under the target policy, whereas the results with a delayed window would only use past data that has a significant state distribution shift. This also explains why methods such as prioritized experience replay \citep{priorizied} can be useful in off-policy settings. \textit{Windowed buffer} experiments suggest that a wide range of samples are necessary to cover as much state space as possible, and in cases with sufficient state space coverage even with smaller windows, using the whole buffer doesn't give any particular advantage. 

\section{Constraining the State Distribution Shift}
\label{sec:constrained_algos}

In this work, we hypothesize that the inability of off-policy deep RL algorithms to learn solely from batch offline data is due to the mismatches in the data distribution under the behaviour and target policies. Since the behaviour policy and updating the target policy based on past experience data can lead to state visitation distribution which is quite different than what would have been the state distribution under the target policy, this mismatch in the state distributions can lead to divergence in off-policy learning algorithms. To remedy this, we propose a trust region KL constraint based on the state distributions under the behaviour policy $d_{\mu}(s)$ and the target policy $d_{\pi}(s)$. Our 
proposed algorithm is based on minimizing the state distribution shift to constrain the target policy updates towards states that are more likely to be visited by the target policy, even though in off-policy RL the samples are drawn from the behaviour policy. In other words, during policy improvement and gradient update, we constrain the target policy parameters $\theta$ towards the target policy state distribution $d_{\pi_{\theta}}(s)$. However, this is often difficult in practice and can lead to sample inefficiency since the target policy is not often accessible in off-policy learning making it difficult to estimate $d_{\pi_{\theta}}(s)$. To this end, we use an approximation where we use the \textit{near on-policy samples} where the samples under the most recent behaviour policy $\mu = \pi_{\theta}(s) + \epsilon$ which is coupled with the target policy and separated by a noise $\epsilon$ (as in algorithms such as DDPG), instead of re-using \textit{any} samples from the experience replay buffer data. We emphasize that most off-policy actor-critic methods do not correct for the state distribution shift under behaviour and target policies \citep{imani2018off,degris2012off,silver2014deterministic,SAC} resulting in the \textit{state distribution shift}. 

\subsection{Estimating the State Distribution}

To constrain the state distribution shift, we present a novel method towards estimating the state distribution based on a density estimator, such that we can still regularize the policy gradient update based on a gradient estimate w.r.t to the state distribution, ie, we can take the gradient $\nabla_{\theta} d_{\pi_{\theta}(s)}$ in the policy gradient update. This is often difficult in practice since we require finding the gradient w.r.t to the state distribution $d_{\pi_{\theta}(s)}$. We highlight that in our approach, we are not estimating the \textit{discounted stationary} state distribution, as we are considering finite horizon episodic tasks, but rather estimating the state density based on the state visitation frequency. We estimate the state distribution based on a density estimator as follows : We use a separately parameterized density estimator with parameters $\phi$, such that we can estimate the log probability density based on the state visitation $\log p(s)$, which we denote by $\log d_{\pi}(s)$. In our approach, we use a generative model such as a variational auto-encoder (VAE) to obtain lower bound approximations to $\log p(s)$. Since we require the log probability density to be dependent on the policy parameters $\theta$, we therefore estimate the state distribution based on state features $\psi_{\pi}(s)$ which is directly influenced by the policy $\pi_{\theta}(s)$. Our model is given as follows : we use a policy network which outputs not just the action probabilities $\pi_{\theta}(a|s)$ but also the state features $\psi(s)$. We can therefore represent the policy network jointly as $\pi(a, \psi(s) | s)$. The state features $\psi_{\pi}(s)$ are given as input to the VAE density estimator, parameterized by $\phi$, which maps the features into a latent space representation $q_{\phi}(z|\psi(s)$ that captures a lower dimensional learnt representation of the state features, and the VAE trained to optimize the following lower bound based on the $KL ( q_{\phi}(Z\mid \psi_{\theta}(S) || p(\psi_{\theta}(S)) )$. 

\begin{equation}
\label{eq:vae_lower_bound}
    \begin{split}
        \log \psi(s)
        &= \log d_{\pi}(S)
        = \log p(S)\\ 
        &>= \E_{q_{\phi}(Z\mid \psi_{\theta}(s)} [\log p_{\phi}(S\mid \psi_{\theta}(s) ]\\
        &- KL ( q_{\phi}(Z\mid \psi_{\theta}(S) || p(\psi_{\theta}(S)) )\\
        &= \mathcal{L}(\phi, \theta)\\
    \end{split}
\end{equation}

In equation \ref{eq:vae_lower_bound} above, the VAE takes as input the state features, instead of the raw states, that are dependent on the policy $\psi_{\pi}(s)$ since it is the output from the policy network $\pi(a|s)$ and directly depends on the states visited by the policy. The encoder in this case learns a latent representation of the state features and the VAE is therefore trained to maximize the log probability of the state features $\log \psi(s)$ instead of the raw states $\log p(s)$. Our estimated state distribution is therefore based on the state features instead of the raw states, which we find to be a more practically feasible approach given the type of benchmark tasks considered for most deep RL algorithms. To maintain clarity in notation, in the text as follows we will denote this as $d_{\pi}(s)$ and not $d_{\pi}(\psi(s))$, even though our model is based on the state features conditioned on the policy. Since the VAE takes as input the state features directly $\psi_{\pi_{\theta}}(s)$ or $\psi_{\mu}(s)$ (where $\mu(a,s)$ is closely related to $\pi_{\theta}(a,s)$ in practice), this ensures that there exists a gradient w.r.t to the ELBO term is dependent on both the parameters $\theta$ and $\phi$. 
In other words, while the VAE is separately parameterized by $\phi$, there also exists a gradient term w.r.t policy parameters $\theta$ for the ELBO term, since $\psi_{\pi}(s)$ is also directly dependent on policy parameters $\theta$. Based on the states under behaviour policy $\mu(a,s)$ which we denote by $s_{\mu}$ and the target policy $\pi(a,s)$ denoted by $s_{\pi}$ we can therefore obtain lower bound approximations to the state distributions under the behaviour policy $d_{\mu}(s)$ and the target policy $d_{\pi}(s)$. This allows us to constrain the policy gradient update with a trust region constraint based on the state distribution mismatch $KL(d_{\mu}(s) || d_{\pi}(s))$. Below we include details about estimating $d_{\mu}(s)$ and $d_{\pi}(s)$ which is key to our proposed algorithm.

\textbf{Estimating $d_{\pi}(s)$ : } To estimate $d_{\pi}(s)$, we require samples that are collected by the target policy $\pi(a,s)$ with which to train the density estimator to obtain $d_{\pi}(s)$. However, in off-policy methods, since we do not explicitly use $\pi(a,s)$ to collect samples, we instead use the online samples that are collected based on the current roll-out. Therefore, in practice, instead of truly estimating $d_{\pi}(s)$, we instead compute an approximation $d_{\pi}(s) + \epsilon(s)$, which we can denote by $d_{\pi + \epsilon}(s)$. We then separately store the current online samples with $\mu(a,s) = \pi_{\theta}(a,s) + \epsilon$ in a separate \textit{online reply buffer} based on the current roll-out. These samples are then used to estimate $d_{\pi}(s)$ by training the VAE density estimator with the online samples to obtain a crude approximation for $d_{\pi}(s) \approx d_{\pi + \epsilon}(s)$. 

\textbf{Estimating $d_{\mu}(s)$ : } To estimate the off-policy state distribution is more convenient, since we can now directly use the samples that are stored in the replay buffer memory. The replay buffer stores all the off-policy samples collected under $\mu(a,s)$ with which we can train the density estimator to obtain $d_{\mu}(s)$.

\subsection{Off-Policy Policy Gradient with Constrained Distribution Shift}

We propose an algorithm that constraints the policy gradient update to account for the mismatch in the state distribution induced by the mismatch between the target and behaviour policies. Our algorithm is based on a constrained KL regularizer on the policy gradient, which instead of constraining the policies between current and old policies (e.g as in TRPO \citep{trpo}, propose a similar trust region update based on the state distributions. We estimate an approximation to the state distribution between the target and behaviour state samples to obtain $d_{\pi}(s)$ and $d_{\mu}(s)$. However, since we cannot exactly compute $d_{\pi}(s)$ as the target policy is not accessible in an off-policy learning setting, we compute a near on-policy state distribution $d_{\pi + \epsilon}(s)$ based on using the most recent samples collected under the behaviour policy. In other words, we estimate $d_{\mu}(s)$ using state samples from the replay buffer, whereas $d_{\pi + \epsilon}(s)$ is estimated based on the current set of trajectories under the behaviour policy. This therefore gives us the following state distribution based trust region constrained off-policy policy gradient objective  where the $\expectation$ is with respect to the behaviour policy samples and we use a regularizer $KL( d_{\mu}(s) || d_{\pi_{\theta}}(s))  )$, such that the overall gradient objective is :      
\begin{equation}
    J(\pi_{\theta}) = \E [ \sum_{t=0}^{\infty} \gamma^{t} r_{t}(s,a) - KL( d_{\mu}(s_{\mu}) || d_{\pi_{\theta} + \epsilon}(s_{\pi}))  ) ]
\end{equation}

where for clarity and explanation, the $\expectation$ is w.r.t off-policy states sampled from the replay buffer denoted by $s_{\mu}$ while we also sample the recent most samples generated by $\mu(a,s) = \pi_{\theta}(a,s) + \epsilon$ from the replay buffer, denoted by $s_{\pi}$. The $\expectation$ samples both $s_{\pi + \epsilon}$ and $s_{\mu}$ based on which we estimate $d_{\mu}(s)$ and $d_{\pi_{\theta} + \epsilon}(s)$. Note that, our approach does not require additional samples compared to the benchmark algorithms and both algorithms would require same amount of samples. Indeed, we simply write the above objective to clarify that the KL regularizer is based on $d_{\mu}(s)$ and $d_{\pi+\epsilon}(s)$ for constraining the state distribution shift. Based on the above objective, we therefore get the following trust region regularized off-policy gradient update that minimizes for the state distribution shift with the $KL ( d_{\mu}(s) || d_{\pi}(s)  )$. 

\begin{equation}
        \nabla_{\theta} J(\theta) = \E_{s \sim d_{\mu}(s), s \sim d_{\pi_{\theta}(s) }}[ \nabla_{\theta} Q^{\pi}(s, \pi_{\theta}(s)) - \nabla_{\theta} KL ( d_{\mu}(s) || d_{{\pi_{\theta}}+\epsilon}(s)  ) ]
\end{equation}

In the above expression, the gradient w.r.t $\theta$ exists for the KL term since we estimate $d_{\pi_{\theta}(s)}$ with a density estimator based on state features $\psi_{\pi_{\theta}(s)}$ that are directly dependent on the policy parameters $\theta$. Therefore, backpropagating the gradients w.r.t to the KL term in the above equation will pass the gradients w.r.t to the policy parameters $\theta$. For clarity and notation, we can further expand the gradient term above by re-writing with separate $\expectation$ terms since only the second KL term depends on the states sampled from $d_{\pi_{\theta} + \epsilon (s)}$

\begin{equation}
\label{eq:off_policy_update_KL}
\nabla_{\theta} J(\theta) = \E_{s \sim d_{\mu}(s)}[ \nabla_{\theta} Q^{\pi}(s, \pi_{\theta}(s)) ] - \E_{s \sim d_{\mu}(s), s \sim d_{\pi}(s)}[ \nabla_{\theta} [ \log d_{\mu}(s) - \log d_{\pi_{\theta}}(s)  ]  ]
\end{equation}

In equation \ref{eq:off_policy_update_KL}, the second term is the KL term, the gradient is w.r.t to $\theta$ for the state distribution $d_{\pi_{\theta}}(s)$. This can be interpreted as a trust region regularizer based on the state distributions.

\textbf{Algorithm : } Our proposed algorithm is outlined in the algorithm box given in Appendix. We use a policy network that outputs both the action distributions and the state features $\pi_{\theta}(a, \psi(s) | s)$. As in any off-policy gradient method, we use a behaviour policy $\mu(a,s)$ which is related or coupled with the target policy, e.g $\mu(a,s) = \pi(a,s) + \epsilon$. We store all the samples in a replay buffer memory, using which the off-policy state distribution $d_{\mu}(s)$ is estimated with a VAE. We use the current online samples with $\mu(a,s)$ to estimate $d_{\pi + \epsilon}$ which we assume to closely approximate $d_{\pi}$ since we assume that the additional $\epsilon$ noise would only change the state visitation distribution by an $\epsilon$ amount. The proposed KL regularizer $KL(d_{\mu}(s) || d_{\pi}(s))$ can then be used with \textit{any} off-policy gradient update such as in DDPG, SAC or TD3 with a $\lambda$ weighted regularizer. By minimizing this KL term, our algorithm can therefore minimize the state distribution shift in off-policy actor-critic algorithms.

\section{Experimental Results}
\label{sec:result}

We first demonstrate our approach in the \textit{growing batch} where the agent keeps storing all instances of past experiences in the replay buffer while it learns. Our experimental analysis is based on a range of continuous control Mujoco tasks \citep{todorov2012mujoco} and we compare our proposed algorithm with standard state-of-the-art off-policy policy based methods including DDPG \citep{lillicrap2015continuous}, TD3 \citep{TD3} and Soft Actor-Critic (SAC) \citep{SAC}. In all our experimental results we compare with the above baselines and add our state distribution KL regularizer with a $\lambda$ weighting to show the significance of minimizing the state distribution shift. Detailed experimental results and ablation studies are also included in the appendix, and we provide code for our algorithm as supplementary. We show that minimizing the state distribution shift can significantly improve performance over baselines. All our experimental results are averaged over $5$ random seeds.

\begin{figure*}[!htb]
    \centering
    \includegraphics[width=.31\textwidth]{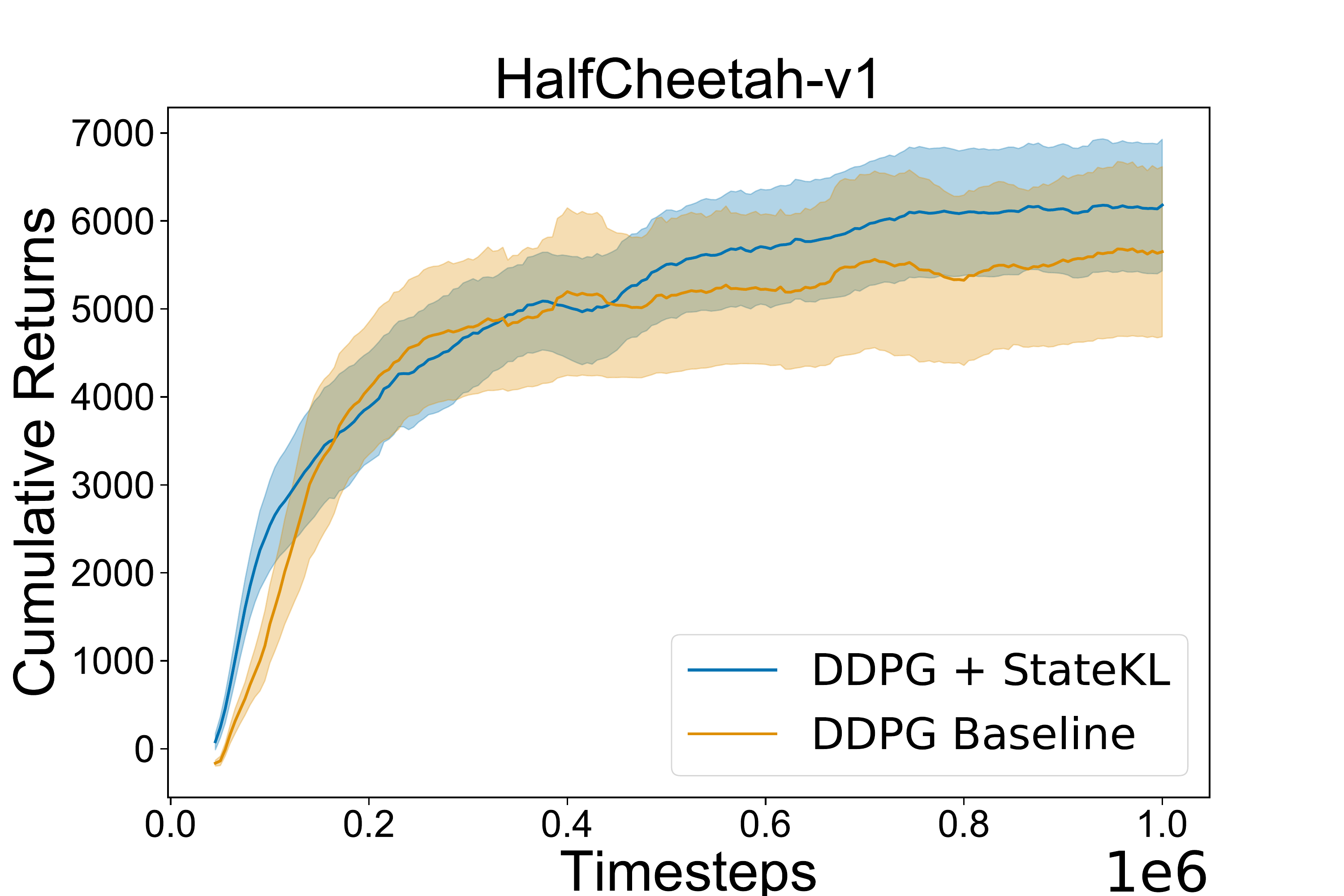}
    \includegraphics[width=.31\textwidth]{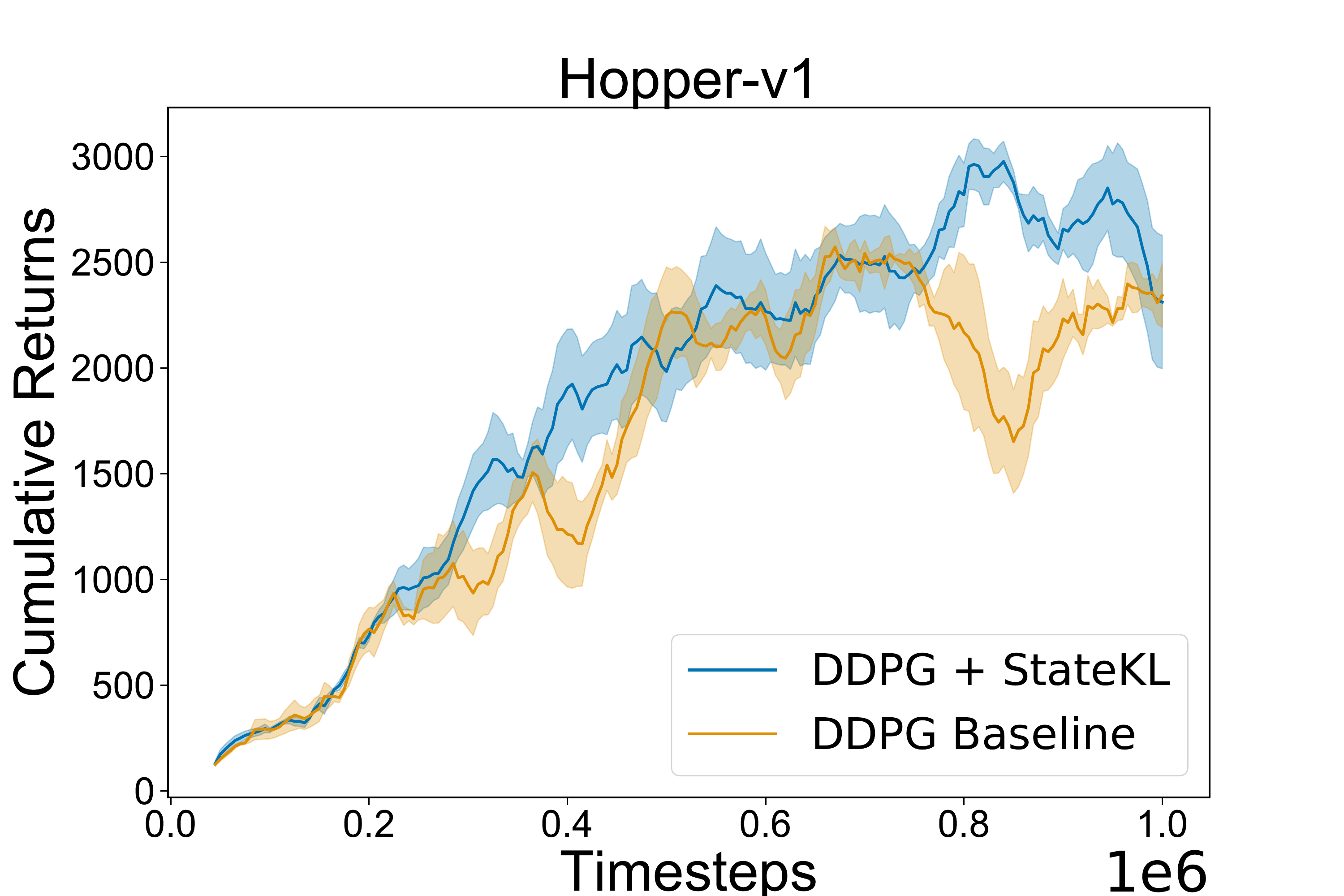}
    \includegraphics[width=.31\textwidth]{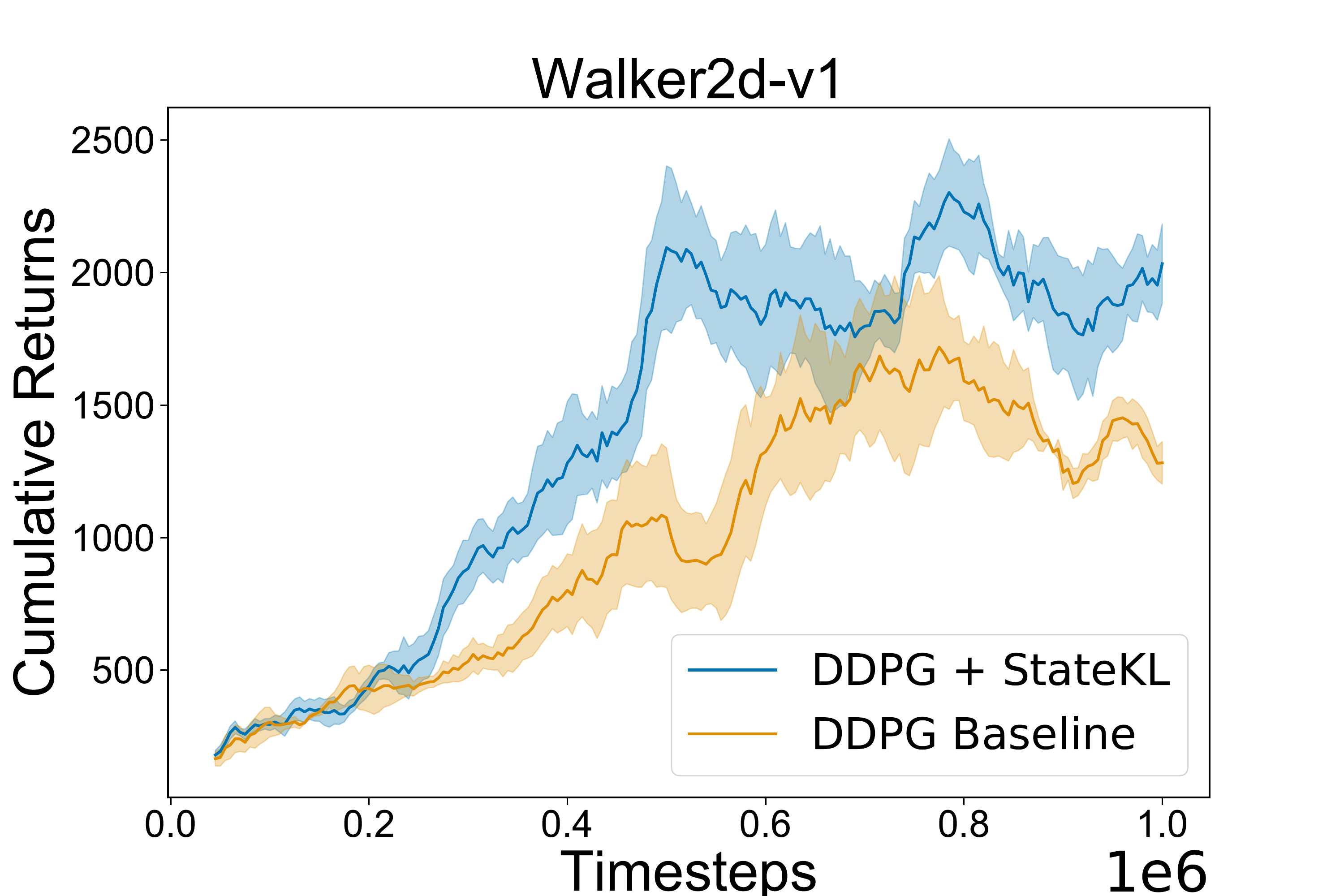}
    \caption{\small{Performance improvements in DDPG algorithm with the proposed "StateKL" regularizer}}
    \label{fig:ddpg_comparison}
\end{figure*}

\textit{Comparisons with DDPG :} We first compare our proposed algorithm with DDPG, where we examine the significance of our proposed approach on several continuous control benchmark tasks. As shown in figure \ref{fig:ddpg_comparison}, we find that with the StateKL regularizer, we can achieve significant performance improvements across all the environments.

\begin{figure*}[!htb]
    \centering
    \includegraphics[width=.31\textwidth]{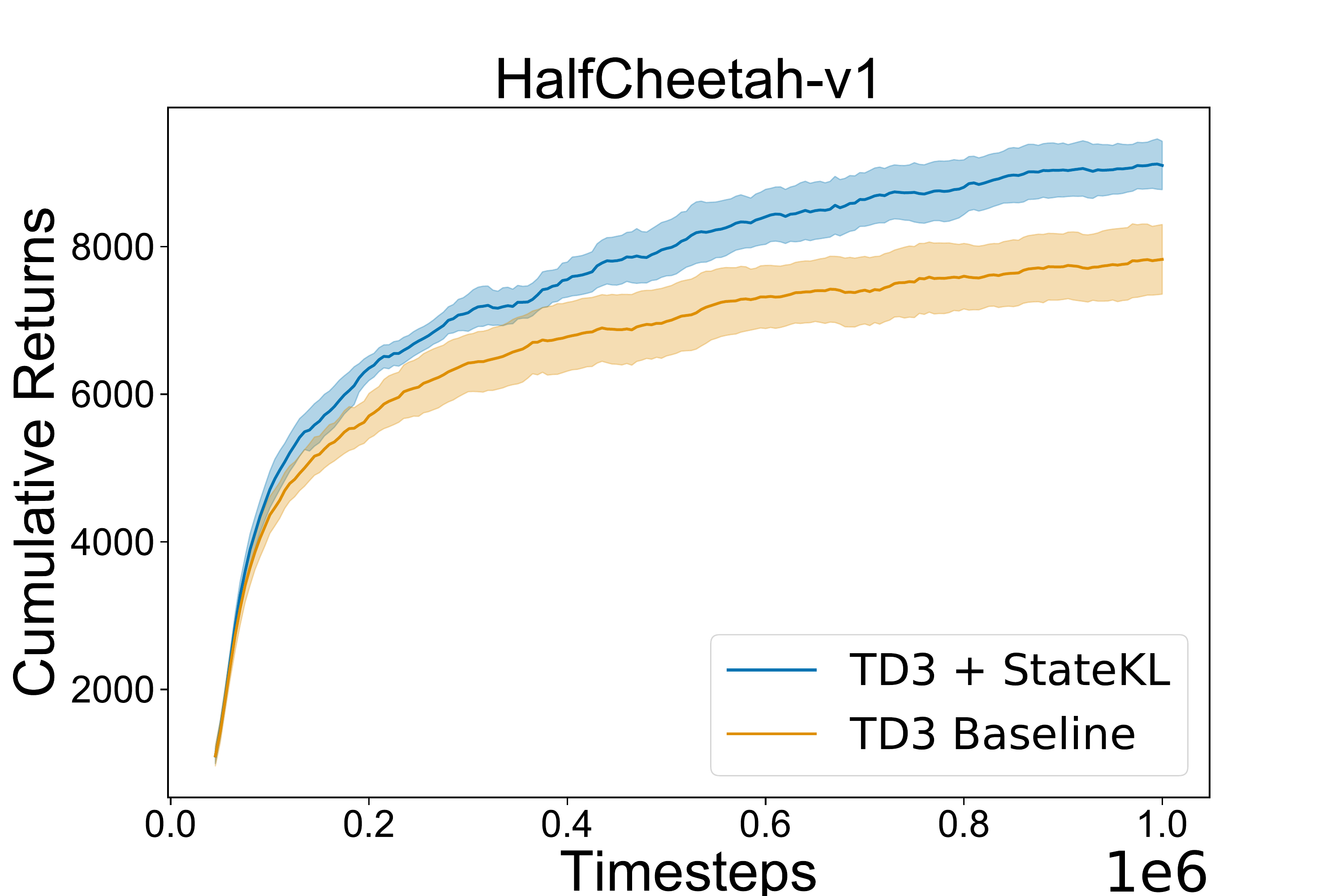}
    \includegraphics[width=.31\textwidth]{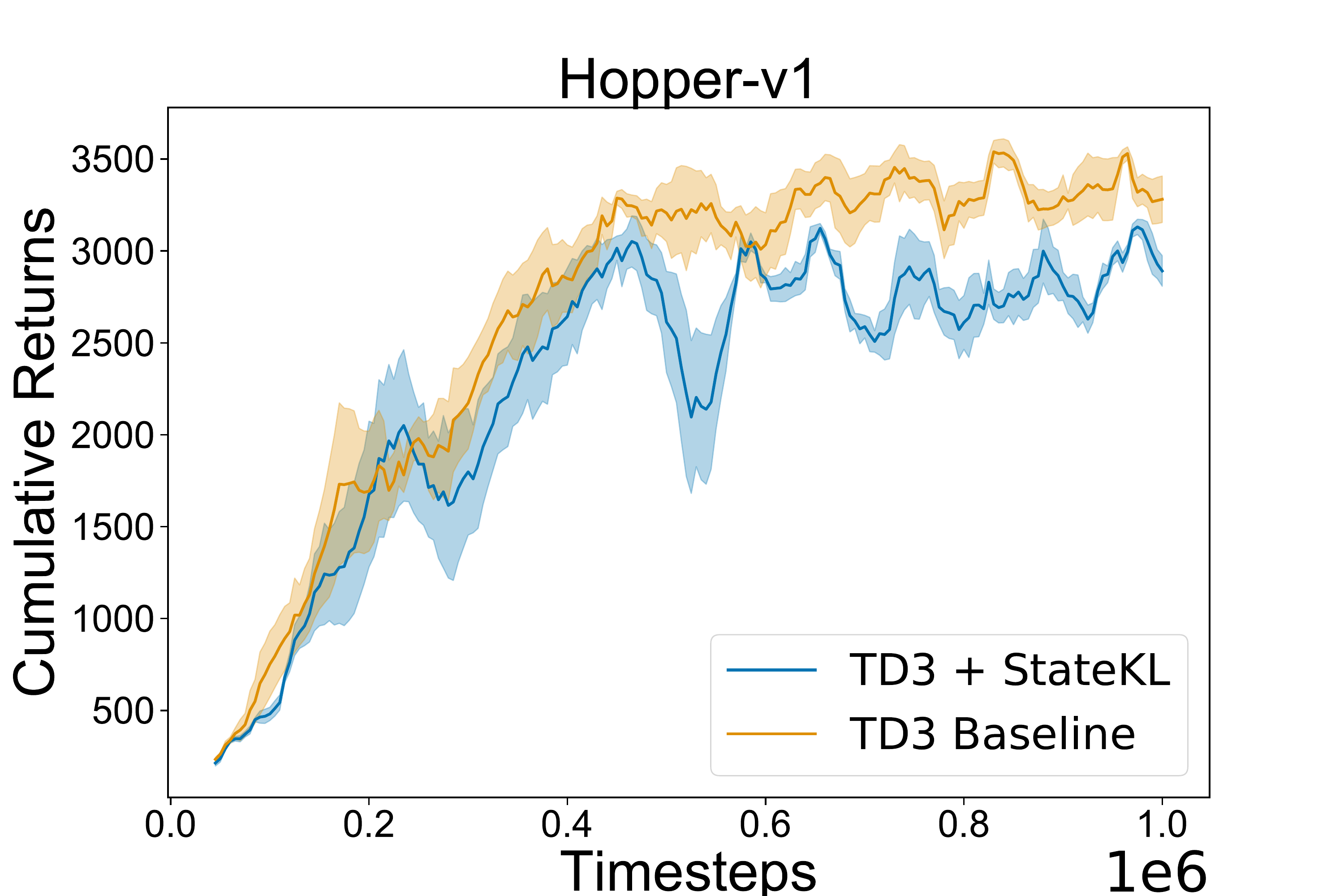}
    \includegraphics[width=.31\textwidth]{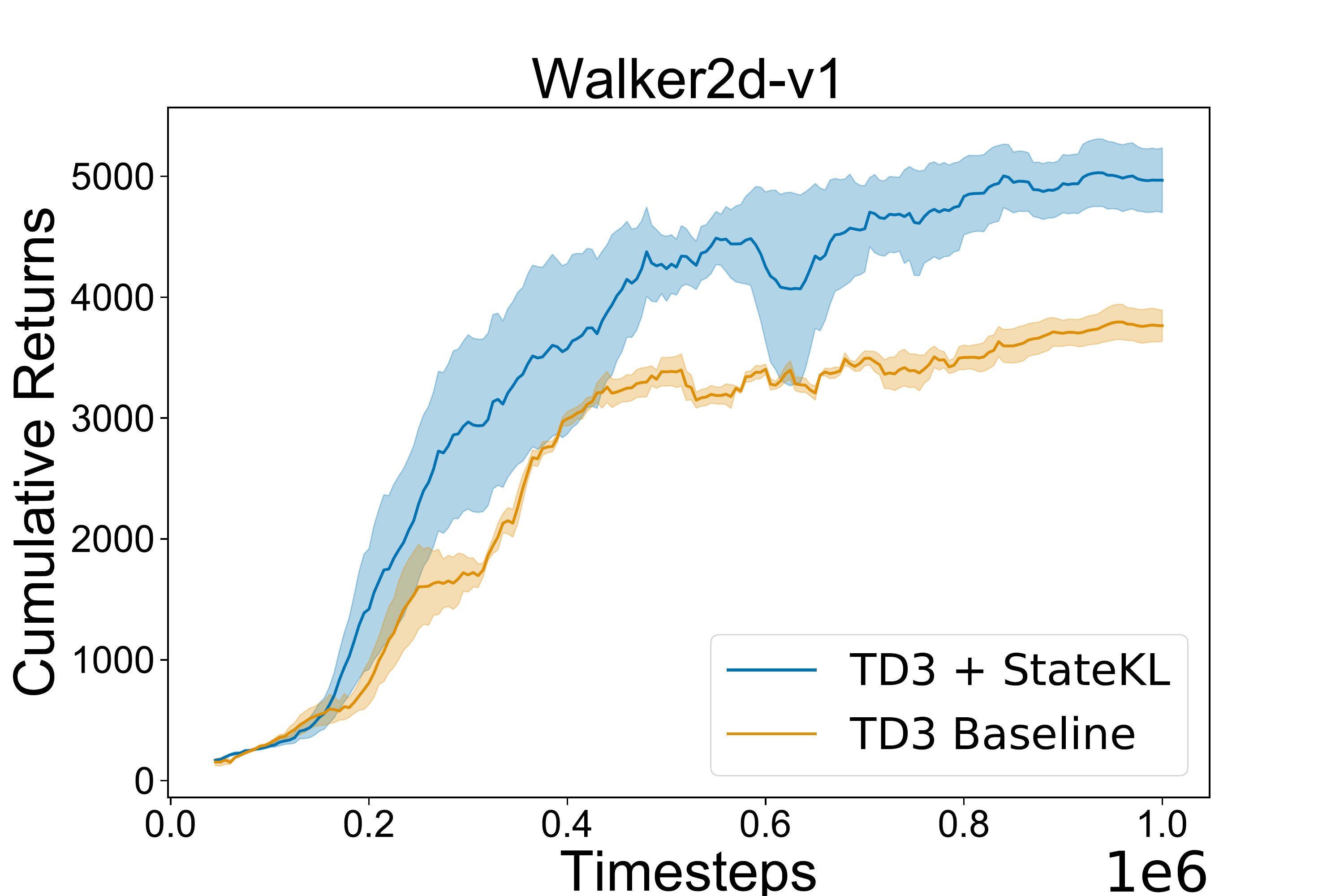}
    \caption{\small{Performance improvements in TD3 algorithm with the proposed "StateKL" regularizer}}
    \label{fig:td3_comparison}
\end{figure*}

\textit{Comparisons with TD3 : } We then analyse the usefulness of minimizing the state distribution shift on TD3 framework as shown in figure \ref{fig:td3_comparison} and find similar improvements in performance across the environments.

\begin{figure*}[!htb]
    \centering
    \includegraphics[width=.31\textwidth]{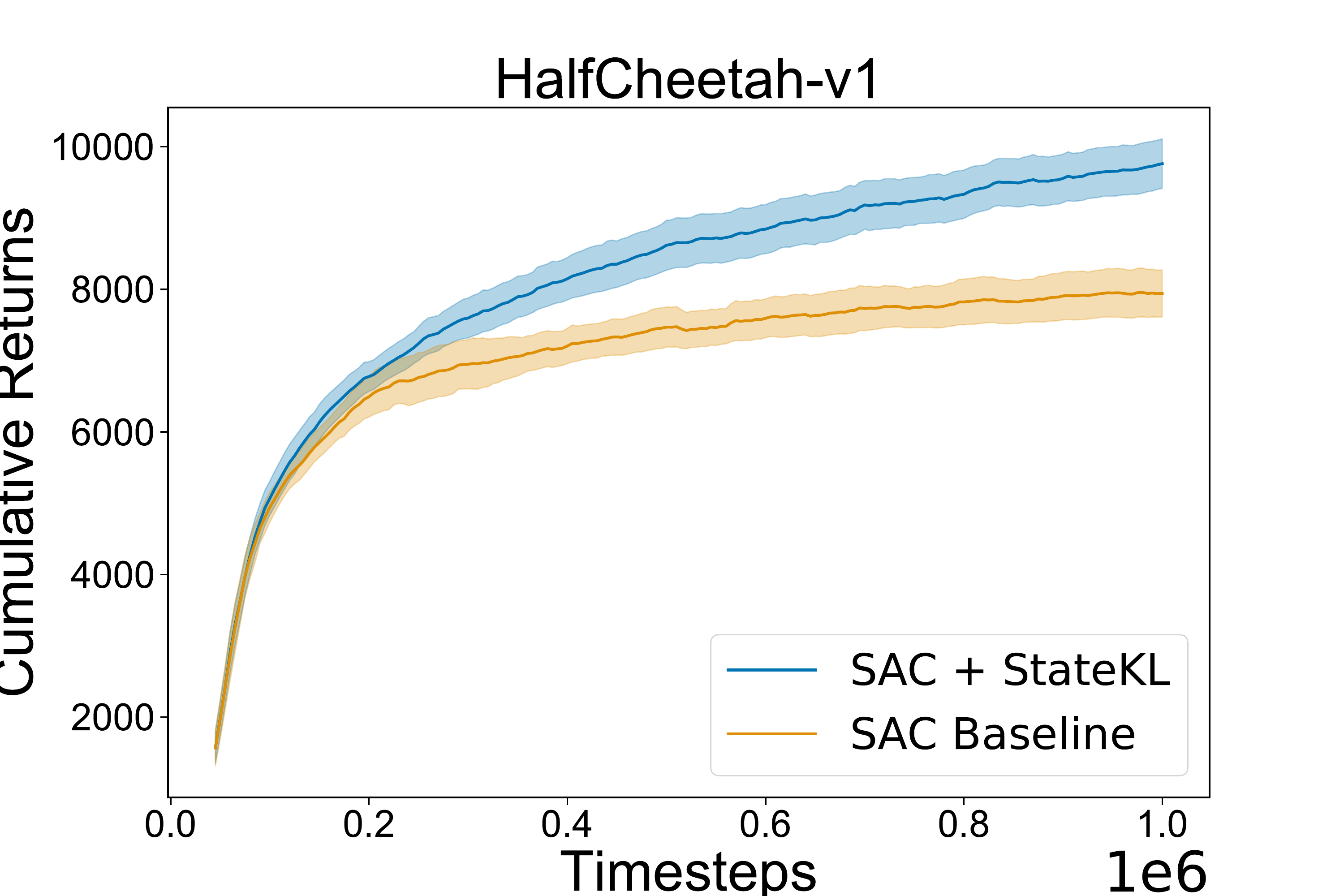}
    \includegraphics[width=.31\textwidth]{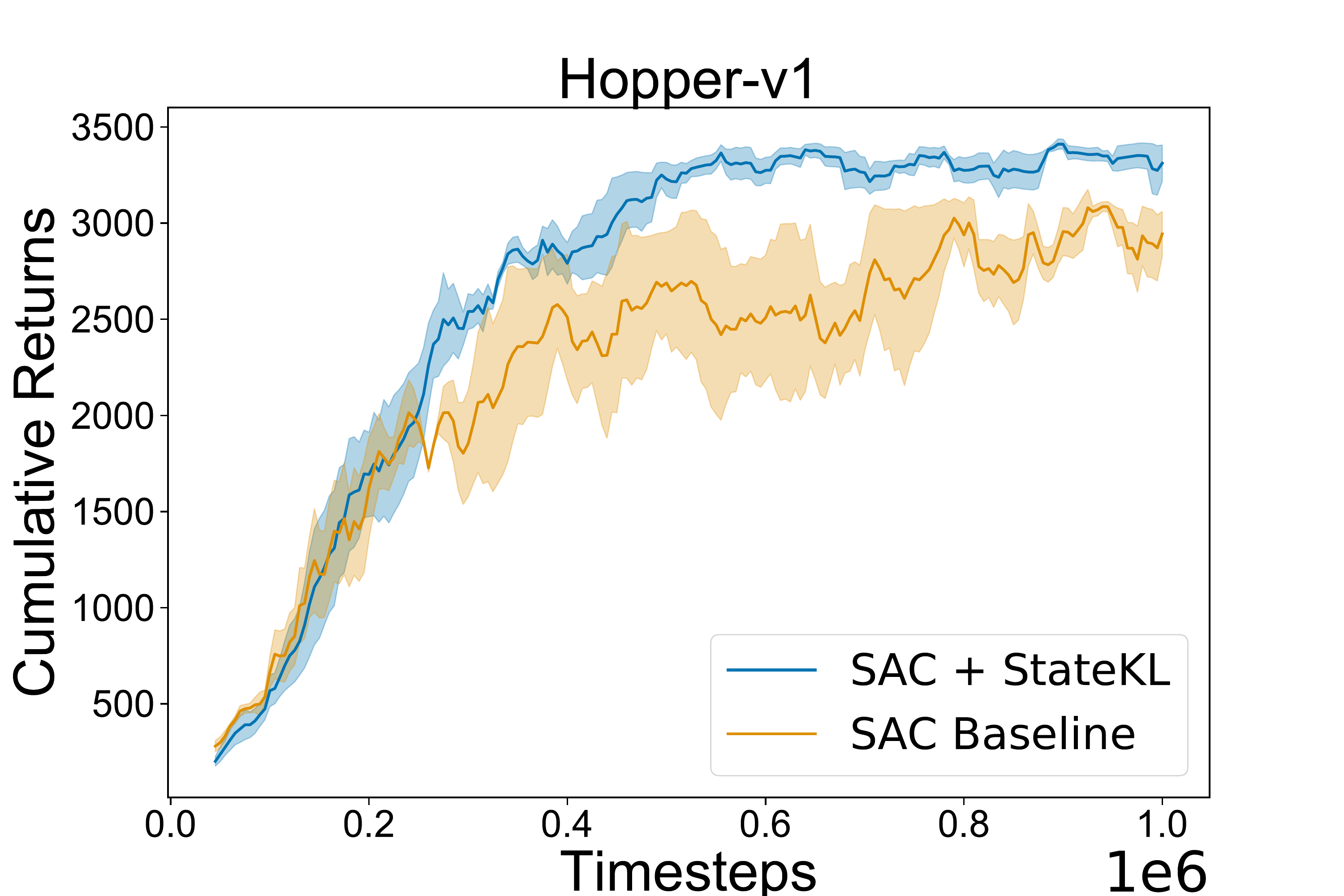}
    \includegraphics[width=.31\textwidth]{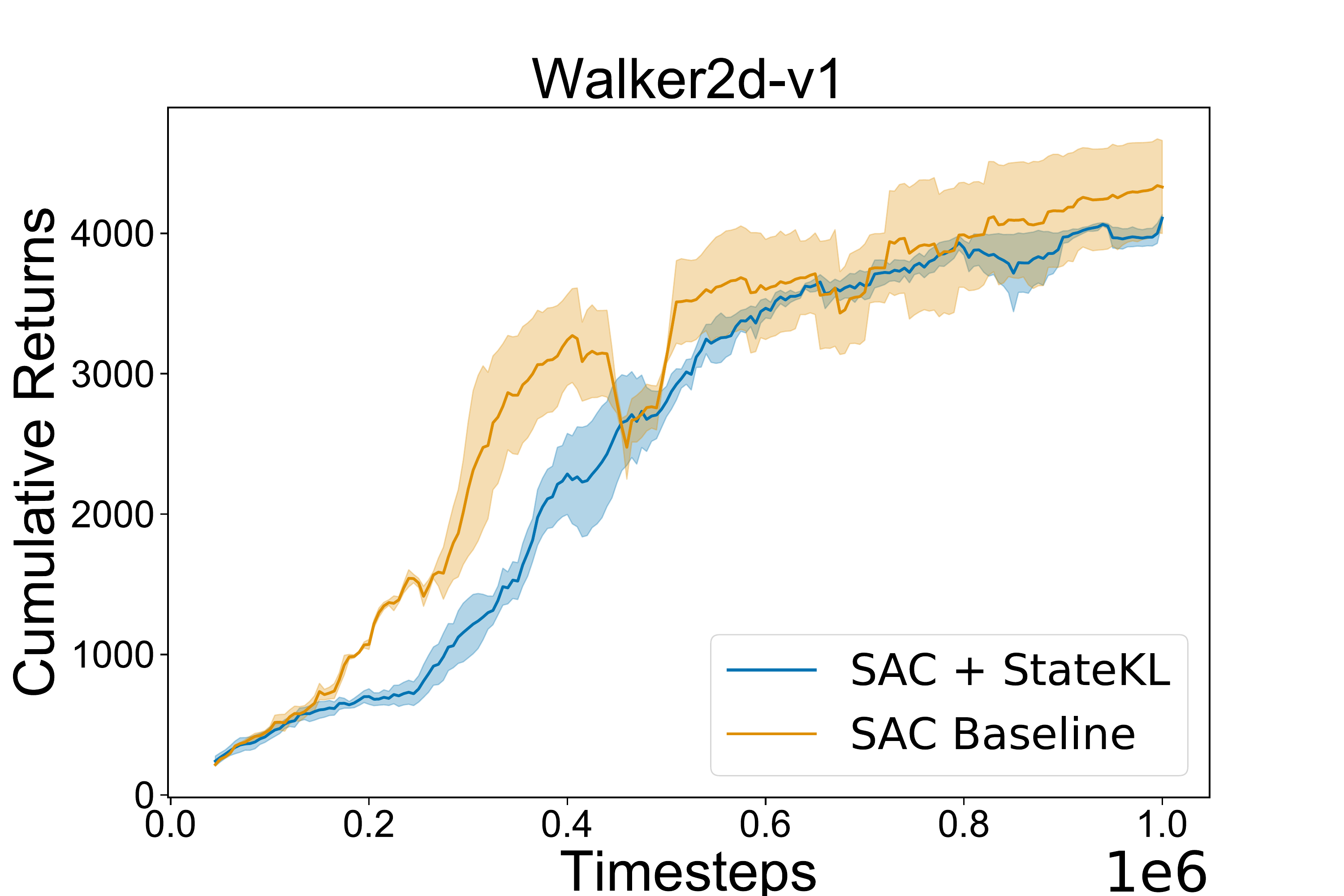}
    \caption{\small{Performance improvements in SAC algorithm with the proposed "StateKL" regularizer}}
    \label{fig:sac_comparison}
\end{figure*}

\textit{Soft Actor-Critic (SAC) : } Finally, we test the significance of our proposed approach on the SAC framework. In two out of the three environments, we find that StateKL regularizer improves the performance over baseline SAC algorithm. 

\textit{Summary : } In all our experimental results comparing with standard off-policy state-of-the-art policy optimization baselines, we find that adding the constraint with the state distribution shift regularizer can significantly improve performance over the baseline algorithms. We note that we do not compare with other baselines here, since our work is the first, to our knowledge, to propose such constraints minimizing the state distribution shift. This is also partly because of our novel framework to estimate the state distribution as well. While other works correct for the mismatch in the state distribution by adding an importance correction $\frac{d_{\pi}}{d_{\mu}}$ \citep{state_correction}, they do not compare in the deep RL continuous control setting, but rather shows simple benchmark tasks making the algorithms incomparable to ours.

\section{Constrained State Distribution Shift in Batch Off-Policy Learning}

We further use our proposed constraint on the state distribution shift in off-policy, in the \textit{pure fixed batch learning} setting \citep{bcqFujimoto2018} with off-policy policy gradient algorithms. As such, the agent is given only a fixed batch of data collected using a pre-trained expert policy, and the goal is to perform policy improvements given this fixed batch of data without further online interactions with the environment. We use the same experimental setup as proposed in the Batch Constrained deep Q-learning (BCQ) algorithm \citep{bcqFujimoto2018}. The BCQ algorithm uses a separate generative model to sample new actions based on data from the batch and use a constrain between the new sampled actions and the actions available in the batch to constrain the policy updates. In our work, we build on the framework of BCQ and use the same setup with additional constraint with the state distribution shift regularizer. We analyze to what extend constraining the state distribution shift $KL(d_{\mu} || d_{\pi_{\theta} + \epsilon})$, in addition to constraining the action space as in BCQ, can help in improving performance of standard off-policy policy based algorithms in the fixed batch setting. As in before, we use all the old instances of the expert data to estimate state distribution $d_{\mu}(s)$, and use the recent most near on-policy samples to estimate $d_{\pi}(s)$ to compute KL. Experimental results below shows the significance of our proposed state KL regularizer for constraining the state distribution shift in batch off-policy learning.

\begin{figure*}[!htb]
    \centering
    \includegraphics[width=.31\textwidth]{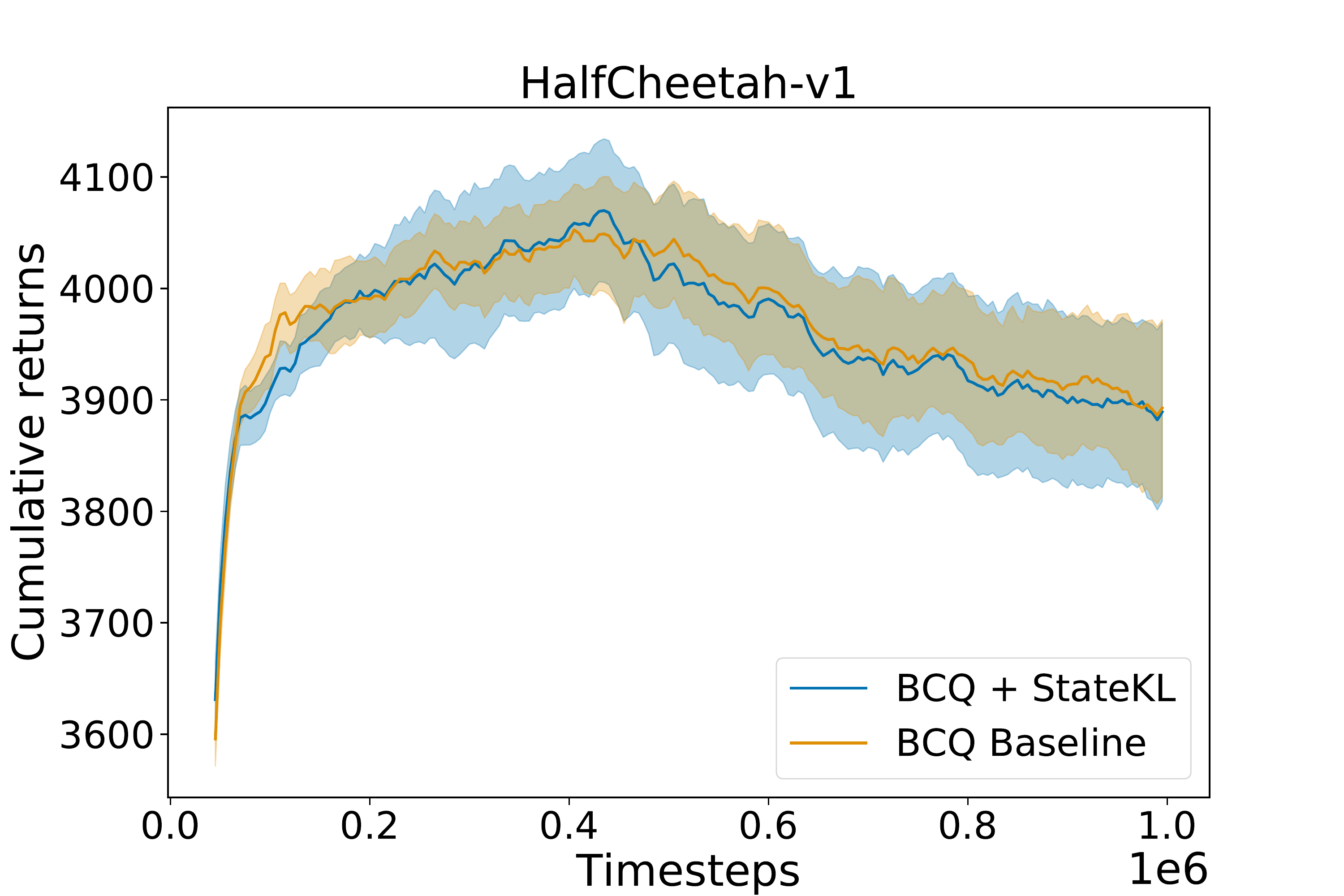}
    \includegraphics[width=.31\textwidth]{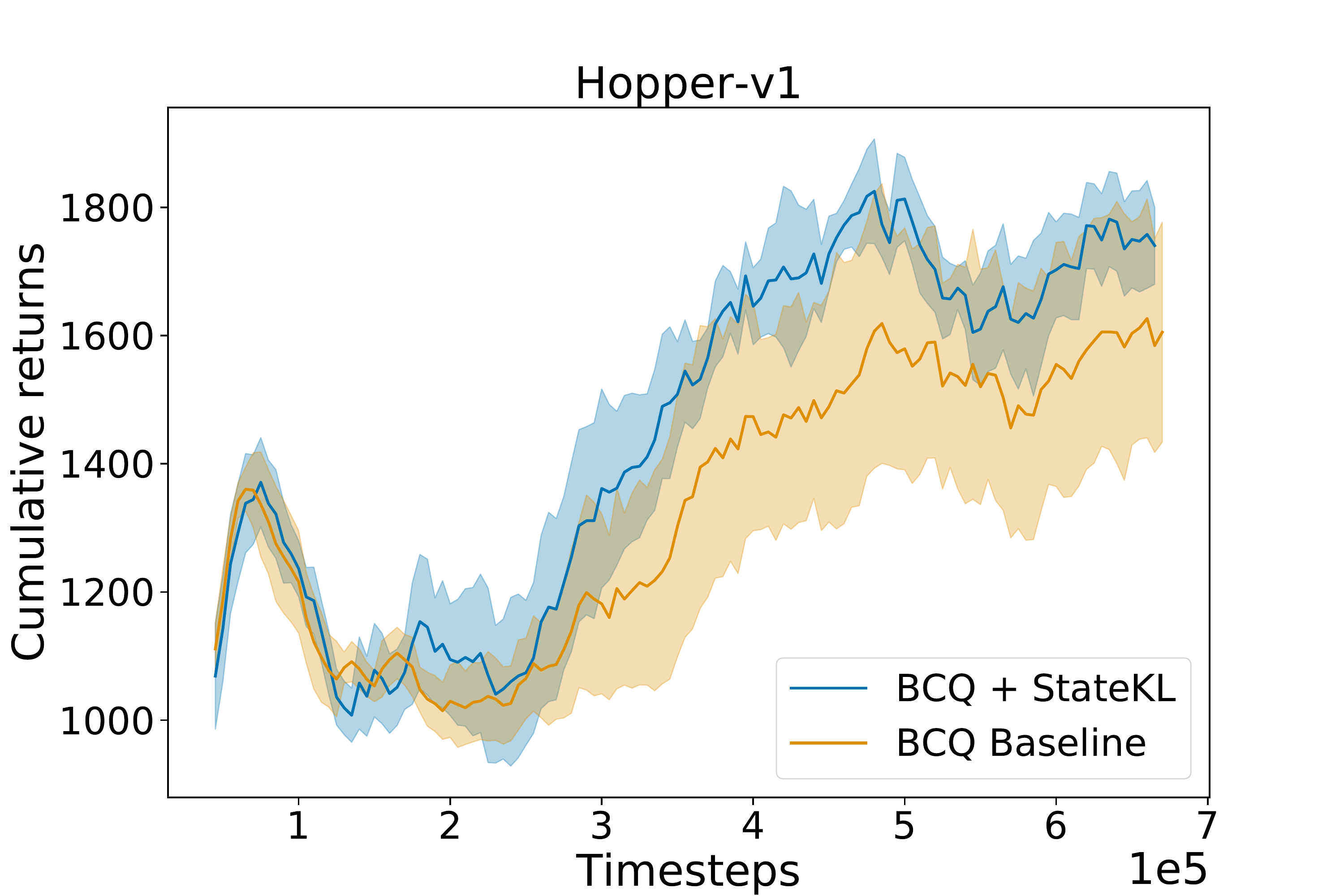}
    \includegraphics[width=.31\textwidth]{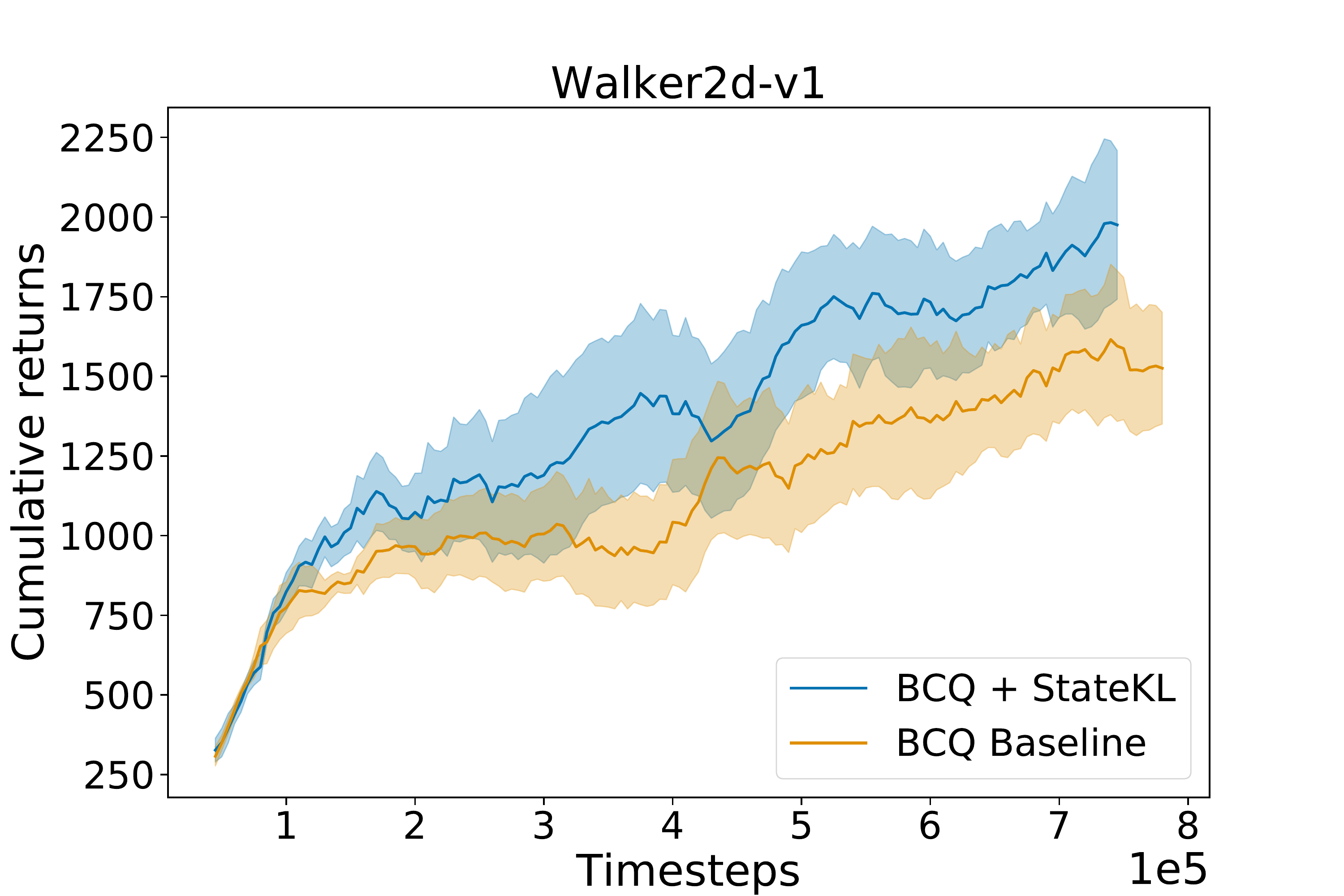}
    \caption{\small{Comparisons with the BCQ Algorithm where we use our proposed KL regularizer for constraining the state distribution shift, in addition to the constrained action space as proposed in the BCQ algorithm for \textit{expert batch data}}}
    \label{fig:bcq_results_fixed_buffer}
\end{figure*}

\begin{figure*}[!htb]
    \centering
    \includegraphics[width=.31\textwidth]{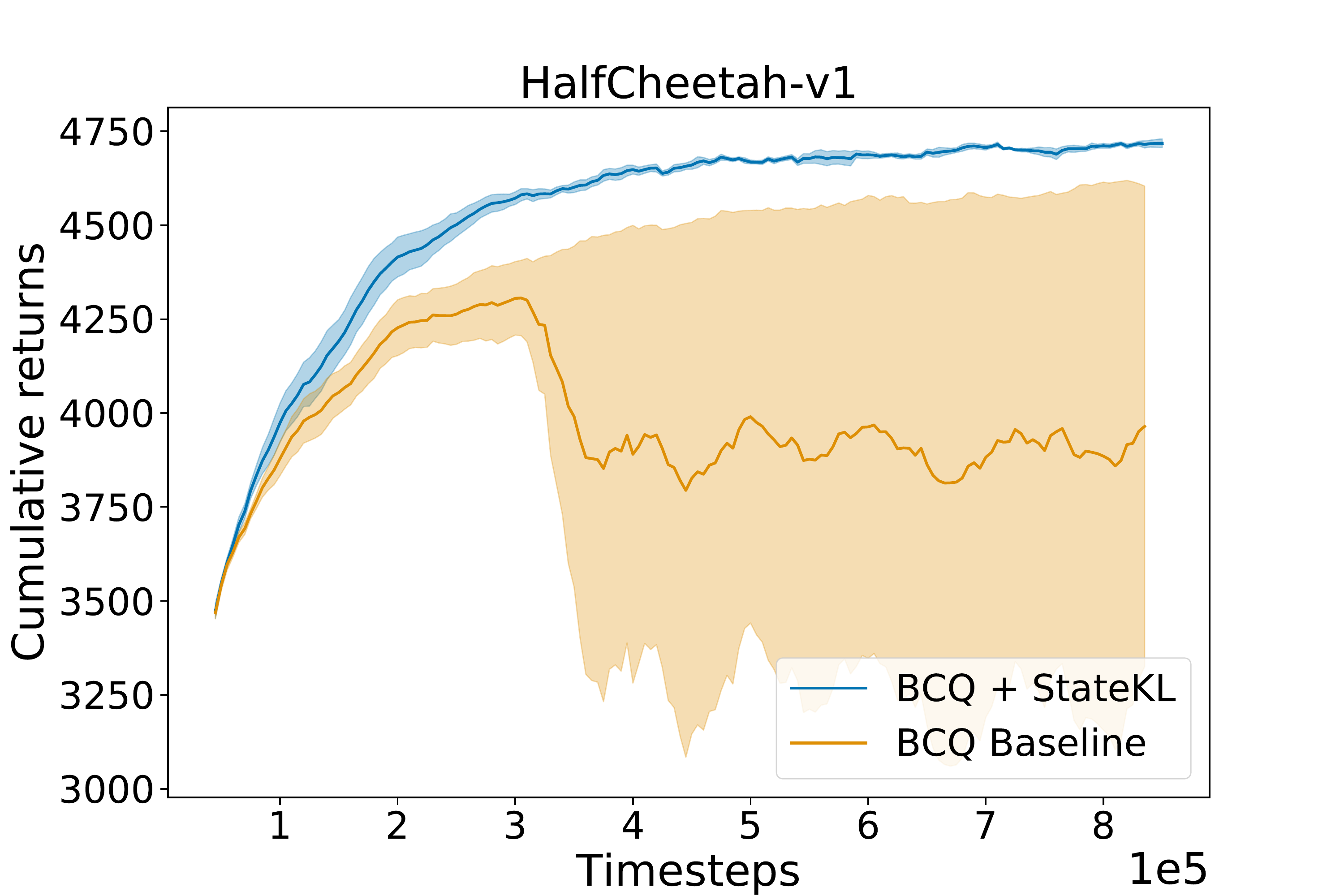}
    \includegraphics[width=.31\textwidth]{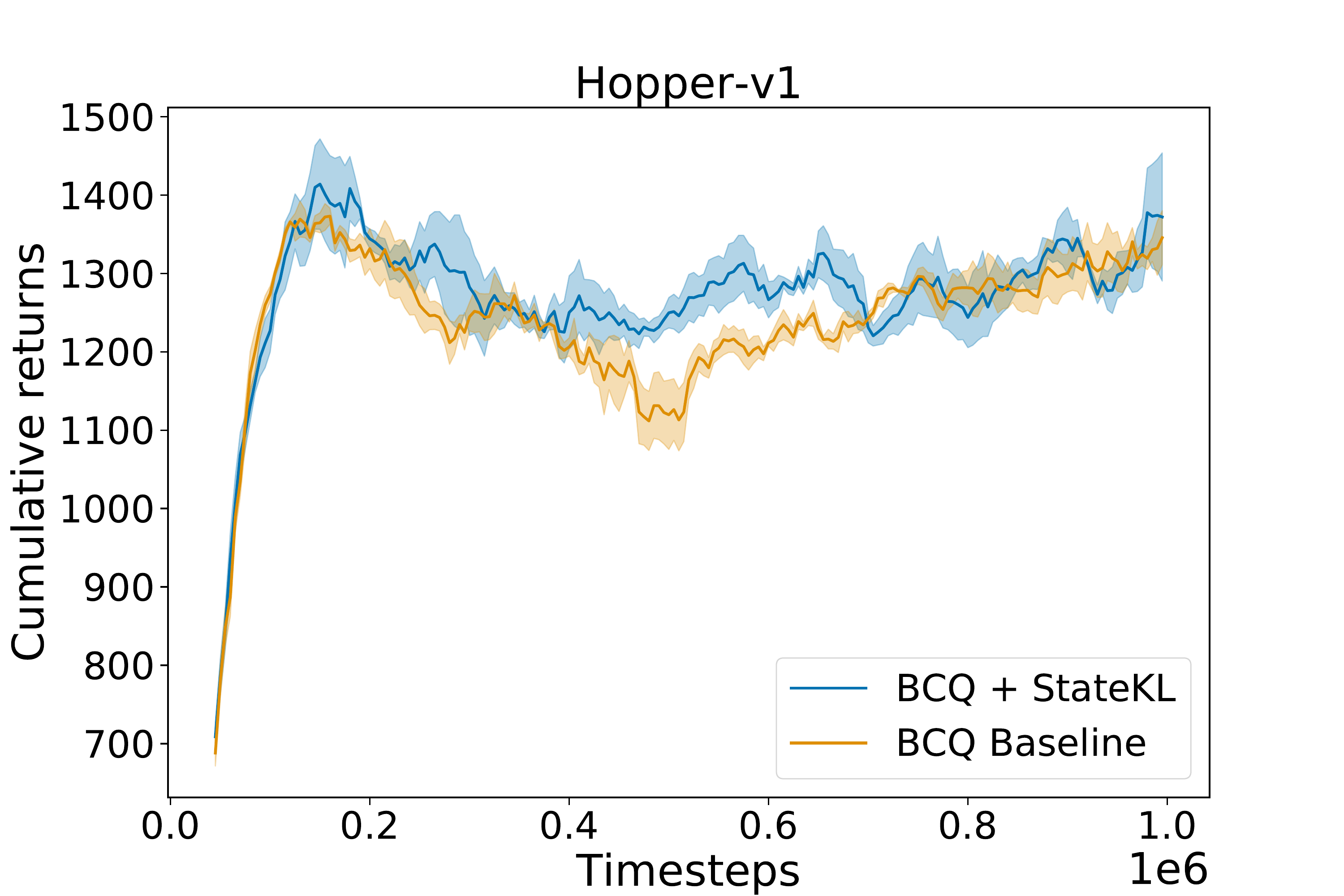}
    \includegraphics[width=.31\textwidth]{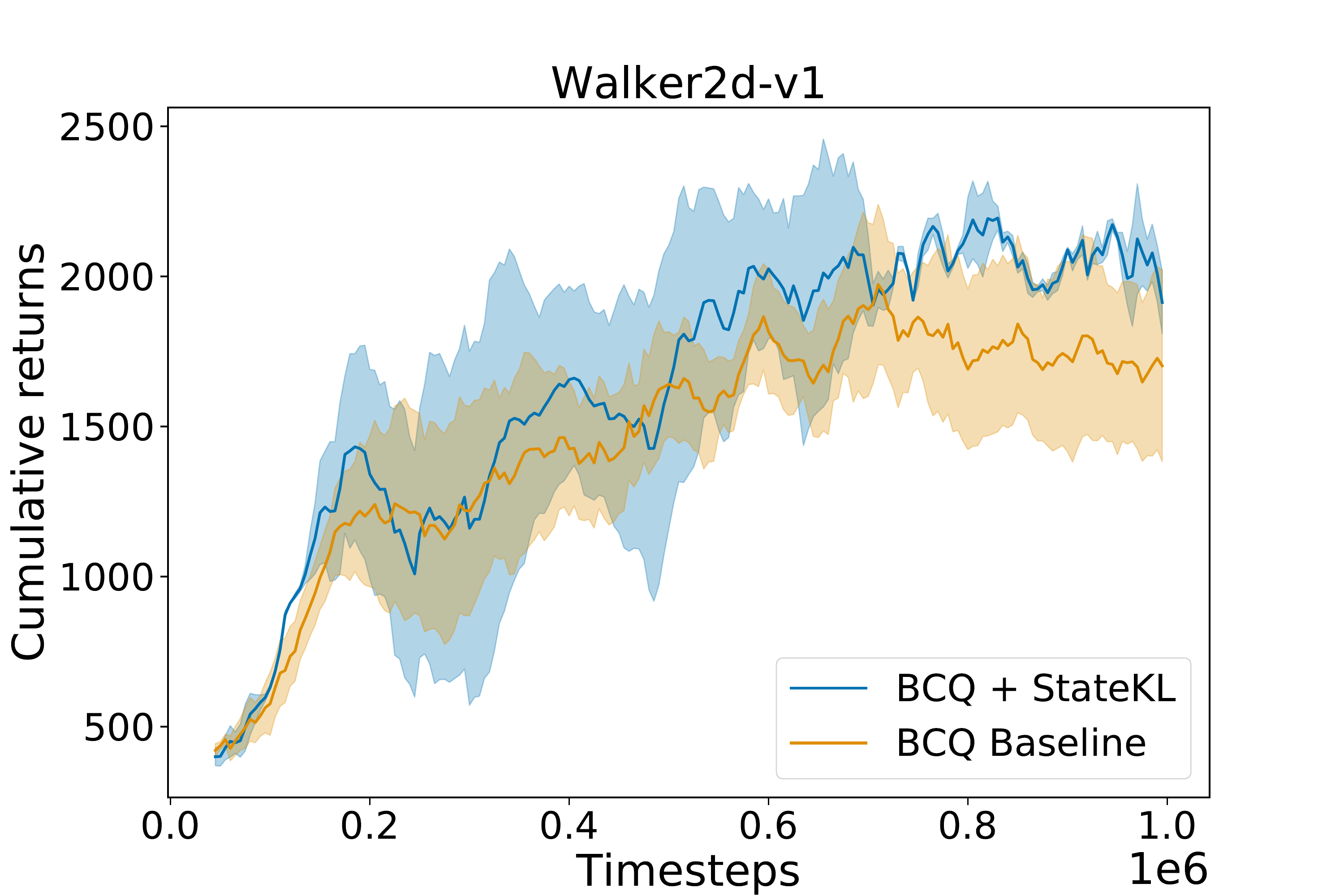}
    \caption{\small{Comparisons with the BCQ Algorithm where we use our proposed KL regularizer for constraining the state distribution shift, in addition to the constrained action space as proposed in the BCQ algorithm for \textit{transient batch data}}}
    \label{fig:bcq_results_transient_buffer}
\end{figure*}

In our experiments, we analyse two different batch settings $(i)$ \textit{Expert batch} : A trained DDPG policy is used to collect the fixed batch of data used to train our agent, and $(ii)$ \textit{Transient batch}: We train a DDPG policy and collect all the transitions experienced during the training process. Figure \ref{fig:bcq_results_fixed_buffer} shows comparisons with BCQ algorithm in the pure batch setting where a given pre-trained expert policy was used to generate the buffer. We show significance of constraining the distribution shift between the near on-policy samples and samples from a very old policy. In figure \ref{fig:bcq_results_transient_buffer} we further compare with BCQ where a growingly trained expert policy was used to collect the buffer. We do this since in this setting it is more meaningful to compute the near on-policy state distribution as we use the recent most expert samples to compute $d_{\pi}(s)$ and old samples from batch to compute $d_{\mu}(s)$.


\section{Discussion and Conclusion}
\label{sec:conclusion}

In this work, we first analyse the ability of off-policy gradient based methods to re-use past samples for data efficiency, and find that off-policy methods are not truly data efficient due to their inability to use \textit{very old} off-policy data. This is because the old samples can have a state distribution which is quite different than the target policy's state distribution, leading to a state distribution shift in off-policy learning. Based on our empirical findings, we then propose a trust region constrained off-policy gradient update which can minimize the state distribution shift between target and behaviour policy state distributions $d_{\pi}$ and $d_{\mu}$. We showed that by estimating the state distribution, based on the state visitation, using a separate density estimator, we can get estimates of the state distribution for both the target $d_\pi(s)$ and the behaviour $d_{\mu}(s)$ policies. We present a novel approach towards estimating the state distribution, which uses an explicit density estimator that takes as input the visited state features under a given policy. Existing off-policy gradient based methods do not correct for the state distribution mismatch, and in this work we show that instead of computing the ratio over state distributions, we can instead minimize the $KL(d_{\mu}(s) || d_\pi(s))$ between the target and behaviour state distributions to account for the state distribution shift in off-policy learning. In our experimental results, we demonstrated the usefulness of minimizing this state distribution shift and showed significant improvements in performance across several continuous control tasks.


\bibliography{neurips_2019}
\bibliographystyle{unsrtnat}

\clearpage

\section{Appendix}

\subsection{Related Work}
Off-policy policy gradient methods are popular for continuous control tasks \citep{lillicrap2015continuous,TD3,SAC} and are mostly derived from the Off-PAC algorithm \citep{degris2012off,imani2018off}. However, none of the previous works accounted for the mismatch in state distributions in off-policy gradient methods. Recently, \cite{state_correction} have introduced an off-policy method that can be used with batch data, while correcting for the difference in the state distributions between current target and behaviour policies. Their approach builds on recent works on off-policy policy evaluation \citep{curse_horizon} by computing a direct ratio over the stationary state distributions under the target and behaviour policies. However, one drawback of their approach is the need to explicitly know the model dynamics in order to compute the stationary distributions. In other words, their approach is not completely model-free and they do not use an explicit estimator of the stationary distributions unlike ours, which makes their algorithm difficult to scale up in practice. Recently, \cite{covariate_shift} also introduced an off-policy correction for the state distributions, where the ratio is directly estimated by bootstrapping. They proposed an algorithm to estimate this ratio directly, based on the covariate shift approach from \cite{COPTD} and extended their results in the deep reinforcement learning framework. In contrast to \cite{covariate_shift} which requires estimating this ratio, where the policy gradient estimator if used with their approach would have high variance in gradient estimators, we instead correct for the mismatch in state distributions by minimizing the state distribution shift based on the KL divergence between the state distributions.

\subsection{Algorithm}

\begin{algorithm}
\caption{Off-Policy Policy Gradient with State Distribution Shift Minimization}
\begin{algorithmic}[1]
\label{algo:training}
\REQUIRE ~~ A policy $\pi_{\theta}(a, \psi(s) \mid s) = \psi(s) p_{\theta}(a \mid s)$ and $\lambda$ regularization coefficient
\REQUIRE ~~ The number of episodes, $E$ and update interval, $N$.
\FOR{$e=1$ to $E$}
\STATE{Take action $a_{t}$,get reward $r_{t}$ and observe next state $s_{t+1}$}
\STATE{Store tuple ($s_t,a_t, r_{t+1}, s_{t+1}$) as trajectory rollouts or in replay buffer $\mathcal{B}$}\\
\STATE{Estimate target policy state distribution $d_{\pi + \epsilon}(s)$ by maximizing variational lower bound (ELBO) using samples collected by $\pi(a,s) + \epsilon$}\\
\STATE{Estimate behaviour policy state distribution $d_{\mu}$ by maximizing variational lower bound (ELBO) using samples collected by $\mu(a,s)$} \\
\STATE{Update policy parameters $\theta$ following \textit{any} policy gradient method, according to\\}
        \begin{equation}
            \begin{split}
                \nabla_{\theta} \tilde{J}(\theta) &= \E_{s_{\mu} \sim d_{\mu}(s), s_{\pi} \sim d_{\pi}(s)}\Bigl[ \nabla_{\theta} Q^{\pi_{\theta}}(s, \pi_{\theta}(s)) \\
                & - \lambda \nabla_{\theta} KL( d_{\mu}(s_{\mu}) || d_{\pi_{\theta}}(s_{\pi}) ) \Bigr] \\
            \end{split}
        \end{equation}
\ENDFOR 
\end{algorithmic}
\end{algorithm}

\subsection{Additional Results : Distribution Shift in Off-Policy Methods }

\begin{figure*}[!htb]
    \centering
    \includegraphics[width=.31\textwidth]{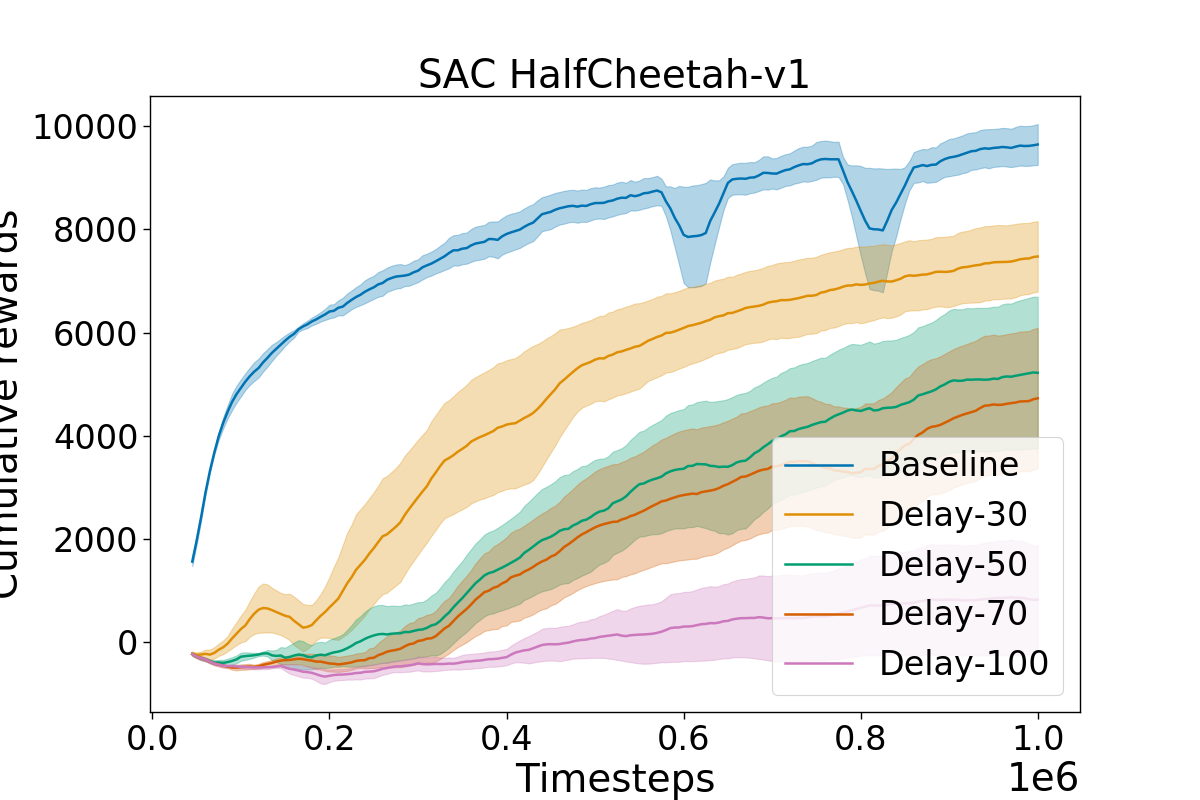}
    \includegraphics[width=.31\textwidth]{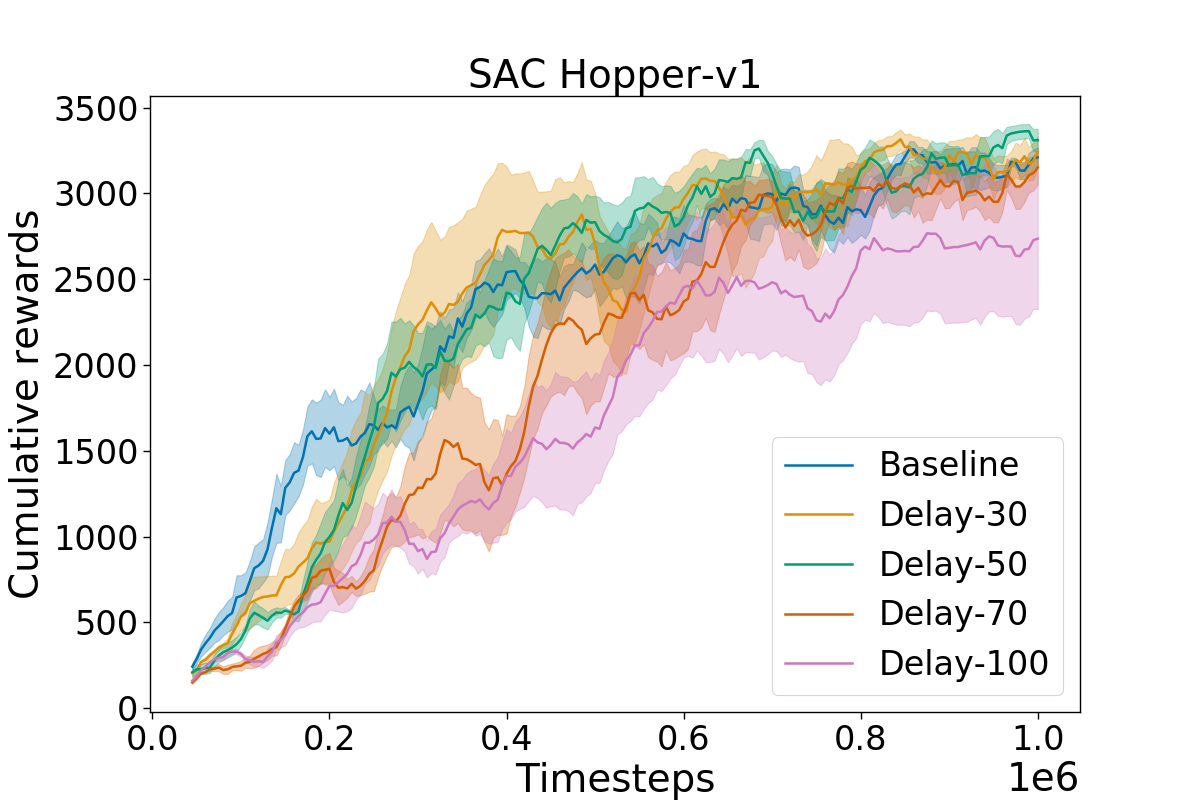}
    \includegraphics[width=.31\textwidth]{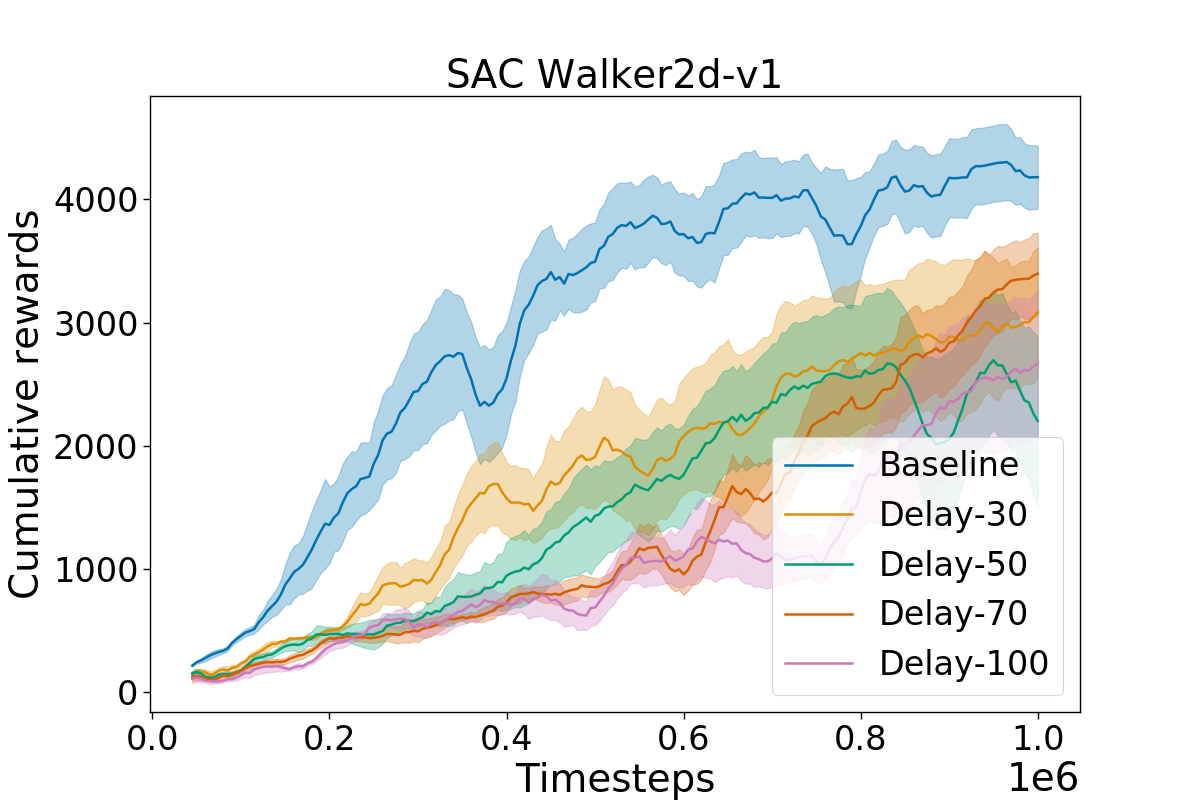}
    \caption{Reward curves for SAC with windowed buffer sampling averaged over \textbf{10} seeds. Windowing is in episodes. The baseline corresponds to window size of $W = 10000$ episodes.}
    \label{fig:sac_delay}
\end{figure*}

\begin{figure*}[!htb]
    \centering
    \includegraphics[width=.31\textwidth]{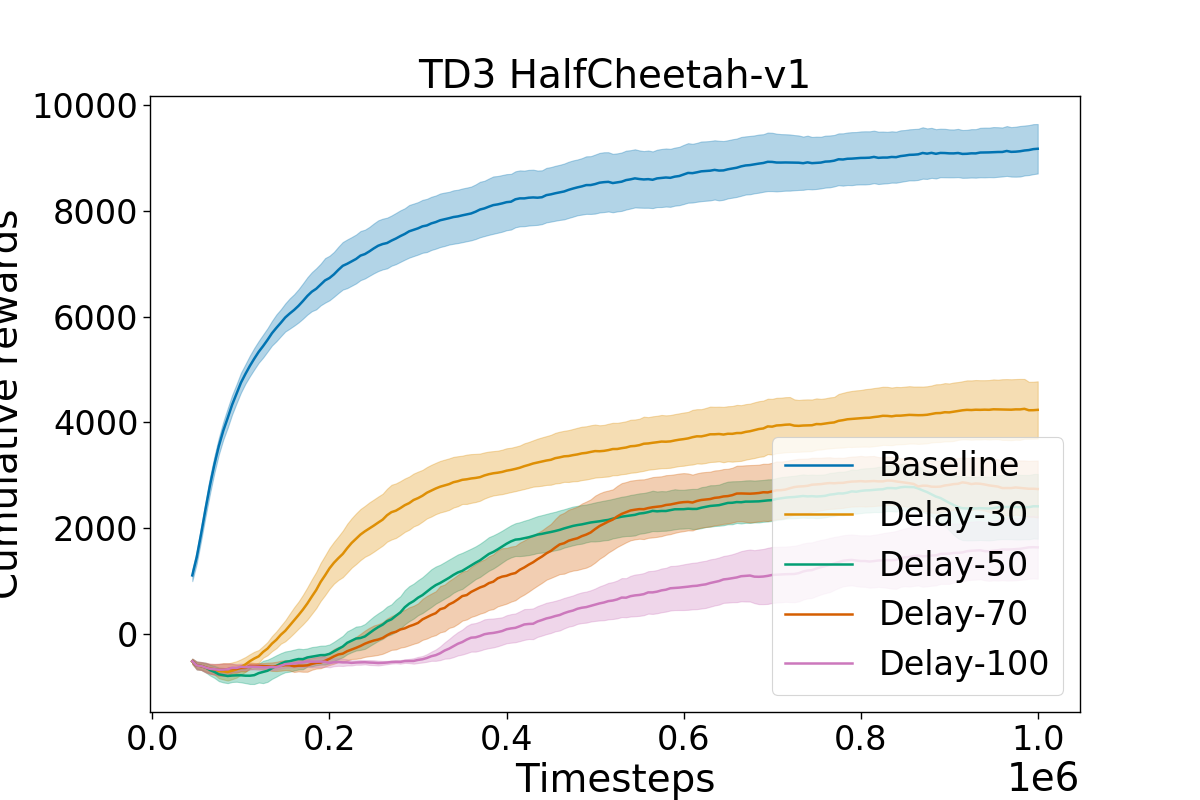}
    \includegraphics[width=.31\textwidth]{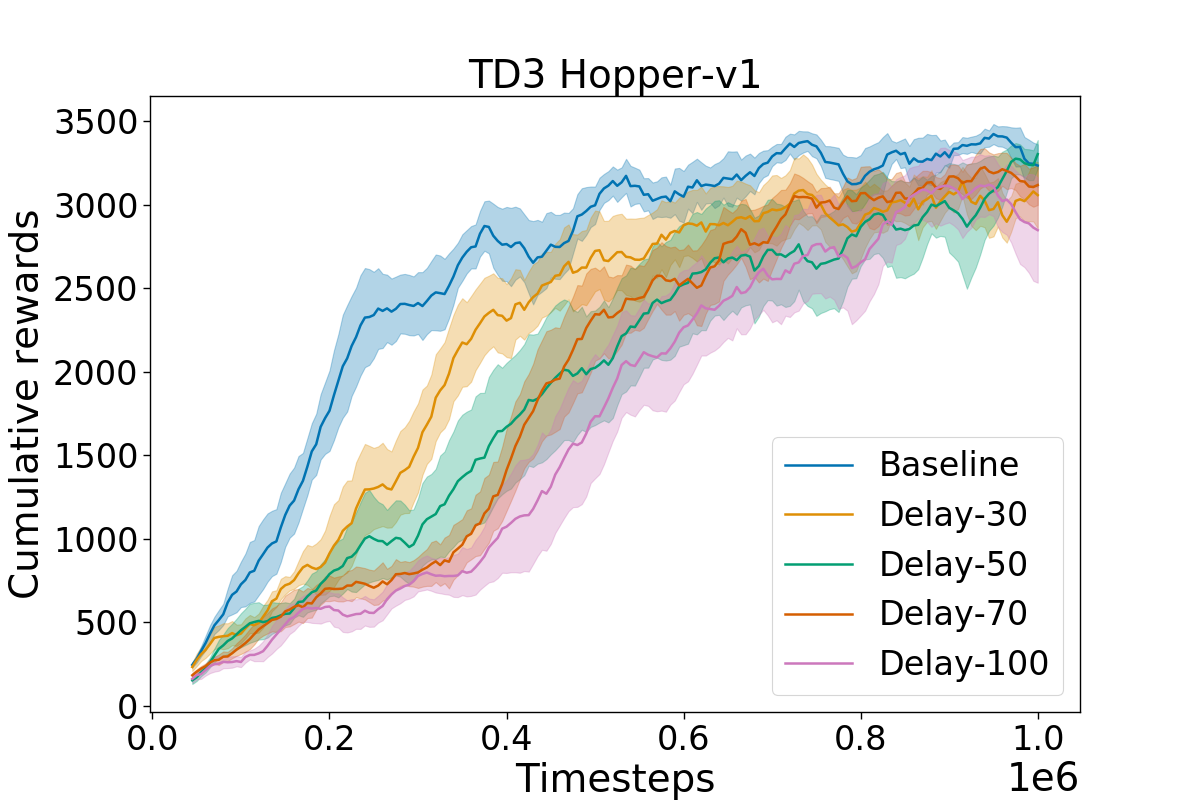}
    \includegraphics[width=.31\textwidth]{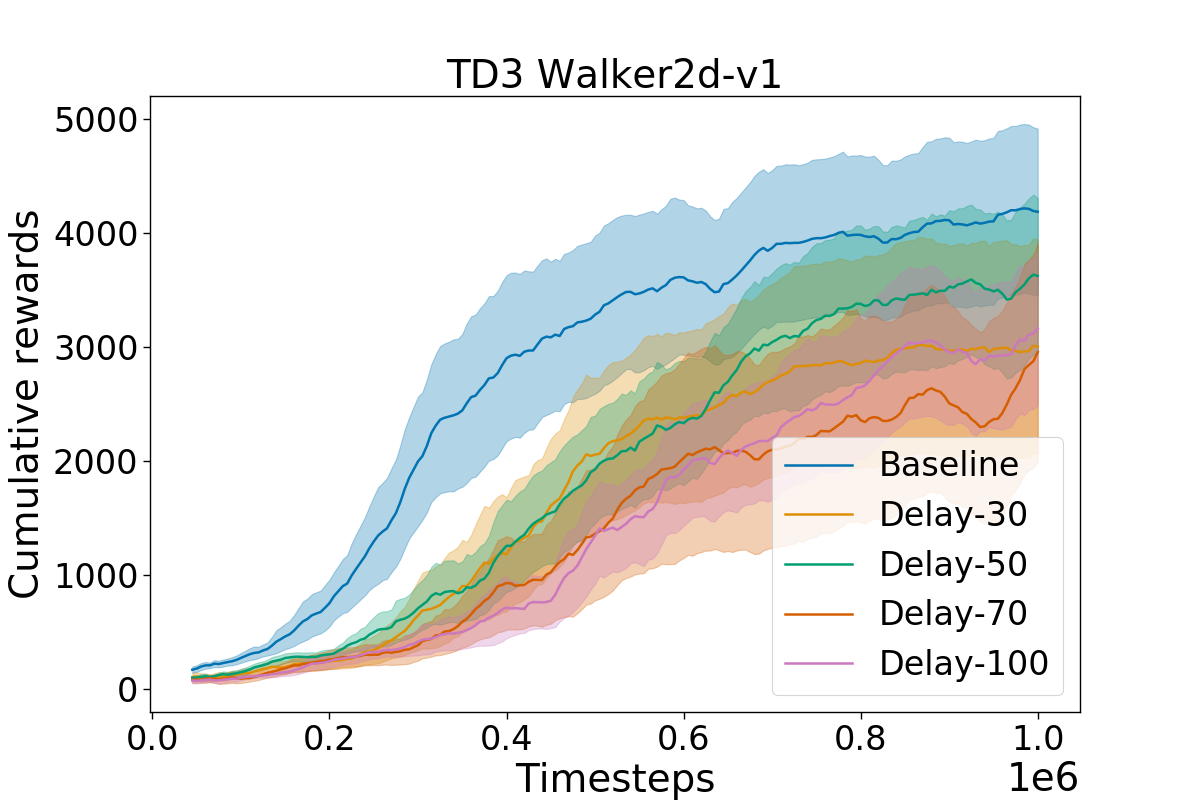}
    \caption{Reward curves for TD3 with windowed buffer sampling averaged over \textbf{10} seeds. Windowing is in episodes. The baseline corresponds to window size of $W = 10000$ episodes.}
    \label{fig:td3_delay}
\end{figure*}

\textbf{Delayed Buffer : } We include additional results here for the \textit{delayed buffer} experiments with both SAC and TD3 in Figure \ref{fig:sac_delay} and \ref{fig:td3_delay}. In the main draft, we demonstrated the impact of using old samples from the replay buffer in DDPG. In the results shown here, we find similar trends in results with both TD3 and SAC as well, where performance drops as we use out of distribution batch samples for training.

\textbf{Summary - Delayed Buffer Experiments : } Our experimental results across all off-policy policy gradient based algorithms therefore conclude that as we increase the distribution shift, by using older samples from the replay buffer, the performance of off-policy policy gradient algorithms degrades. This validates our hypothesis that distribution shift impacts the performance in off-policy learning.

\begin{figure*}[!htb]
    \centering
    \includegraphics[width=.31\textwidth]{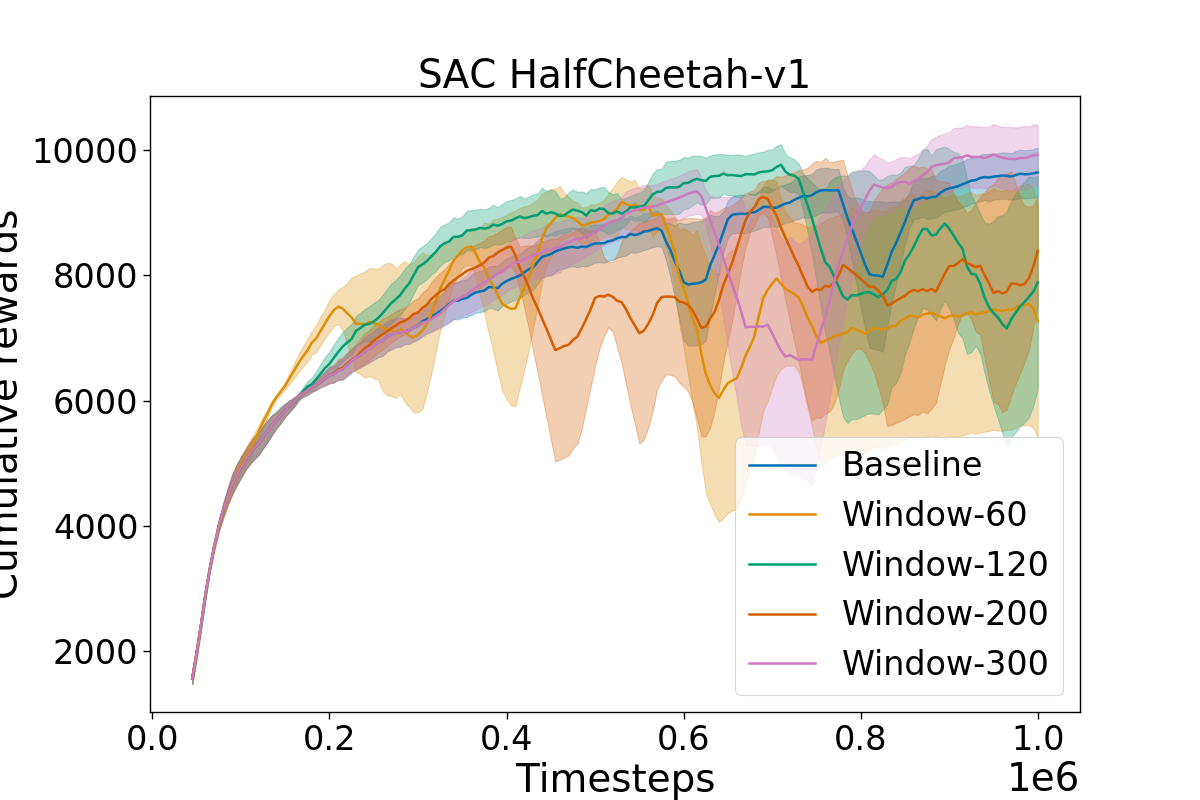}
    \includegraphics[width=.31\textwidth]{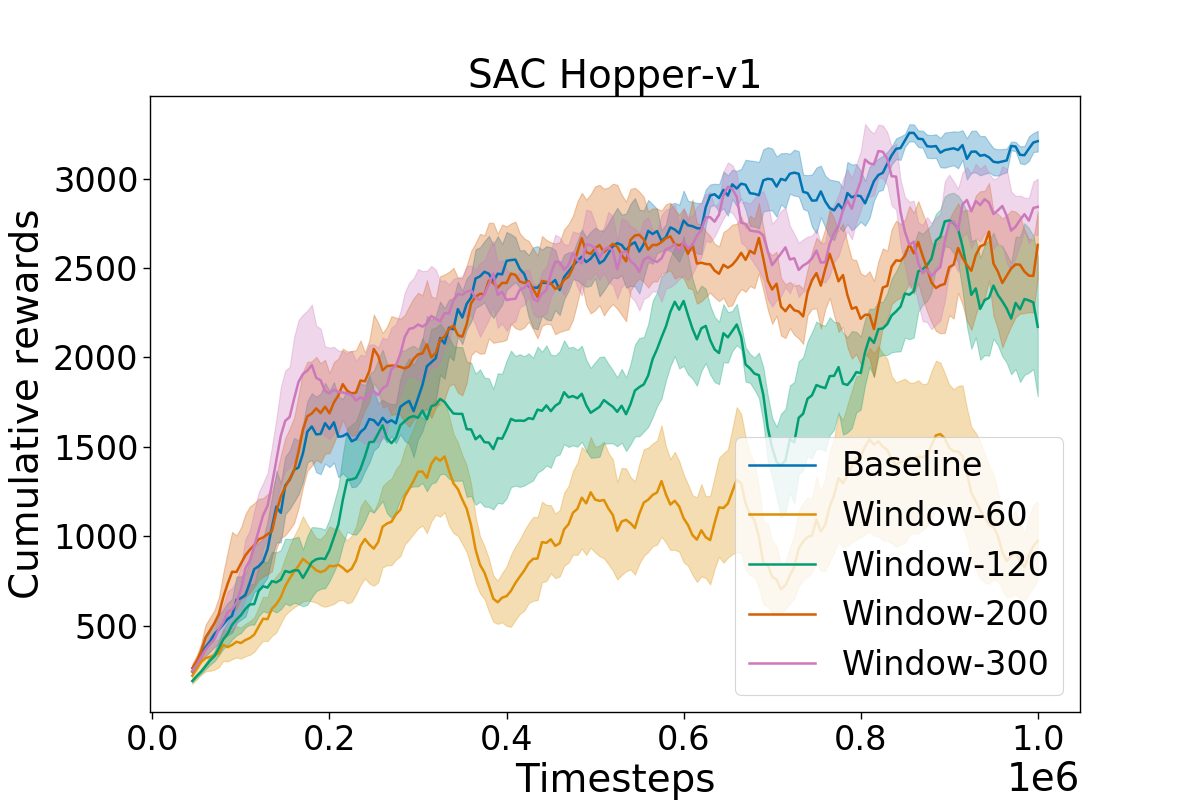}
    \includegraphics[width=.31\textwidth]{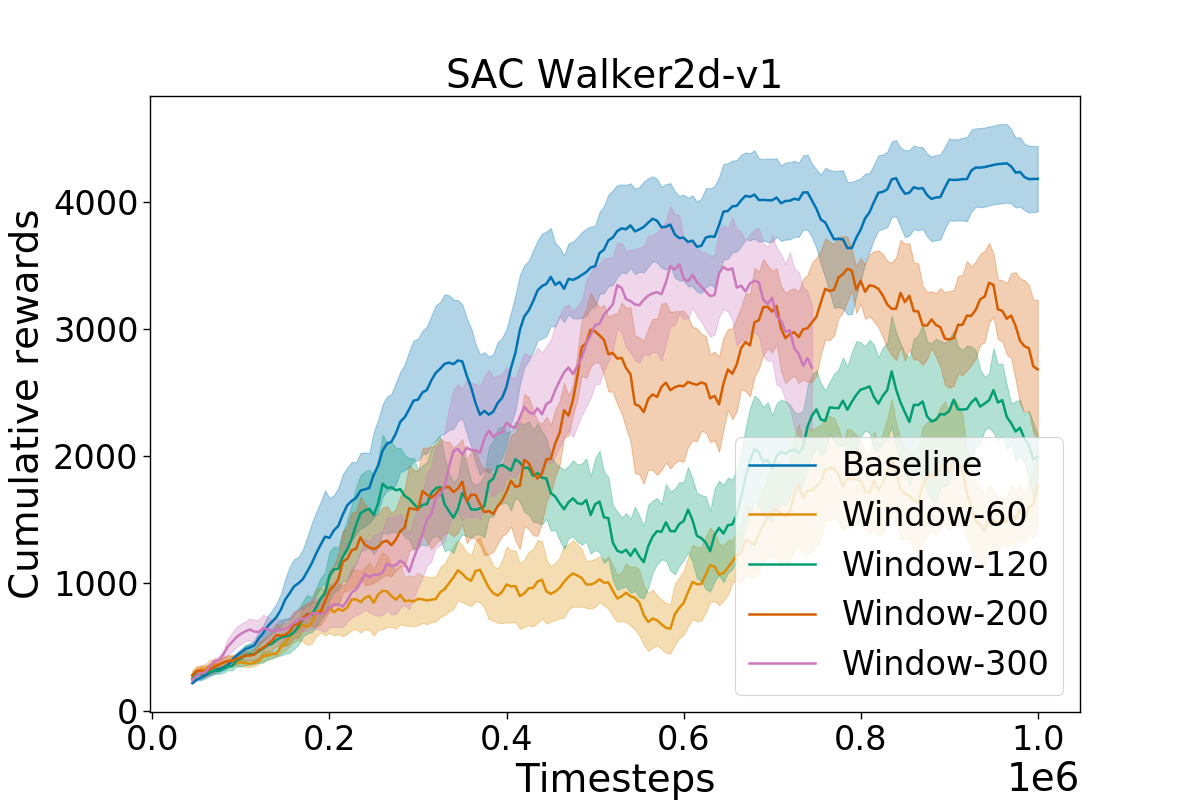}
    \caption{Reward curves for SAC with windowed buffer sampling averaged over \textbf{10} seeds. Windowing is in episodes. The baseline corresponds to window size of $W = 10000$ episodes.}
    \label{fig:sac_window}
\end{figure*}

\begin{figure*}[!htb]
    \centering
    \includegraphics[width=.31\textwidth]{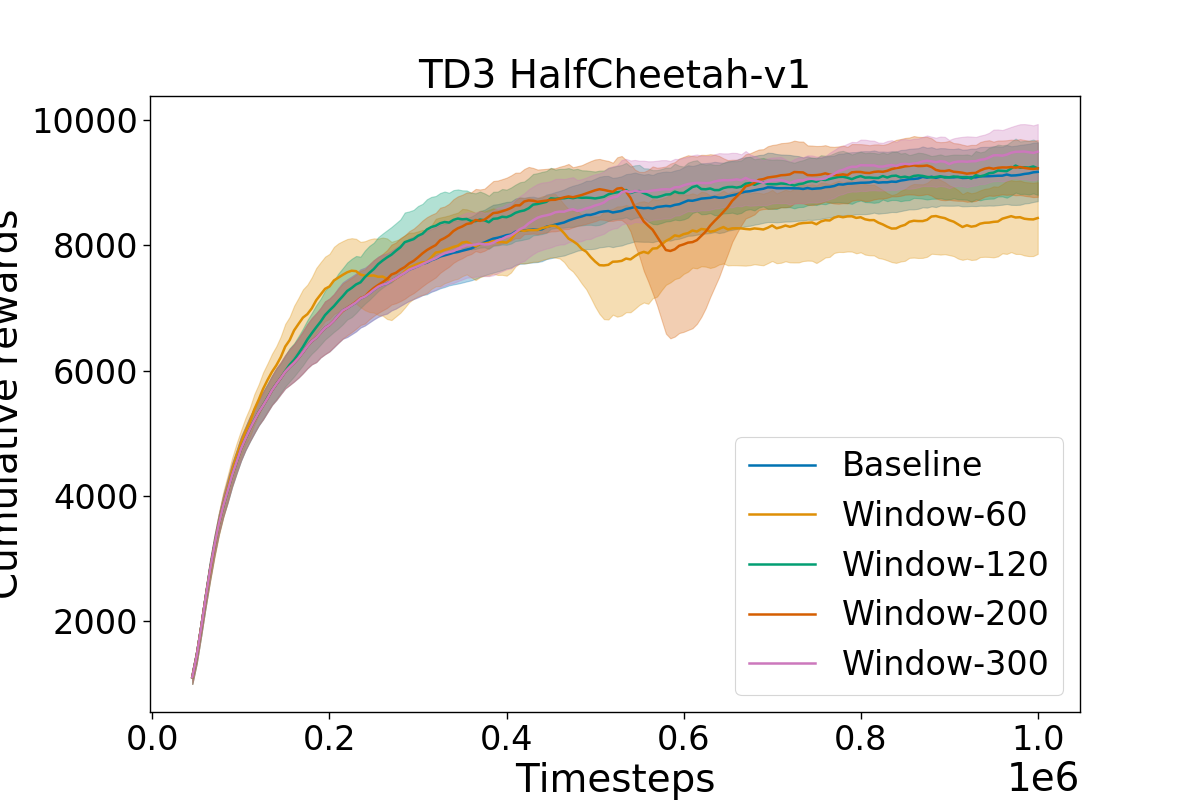}
    \includegraphics[width=.31\textwidth]{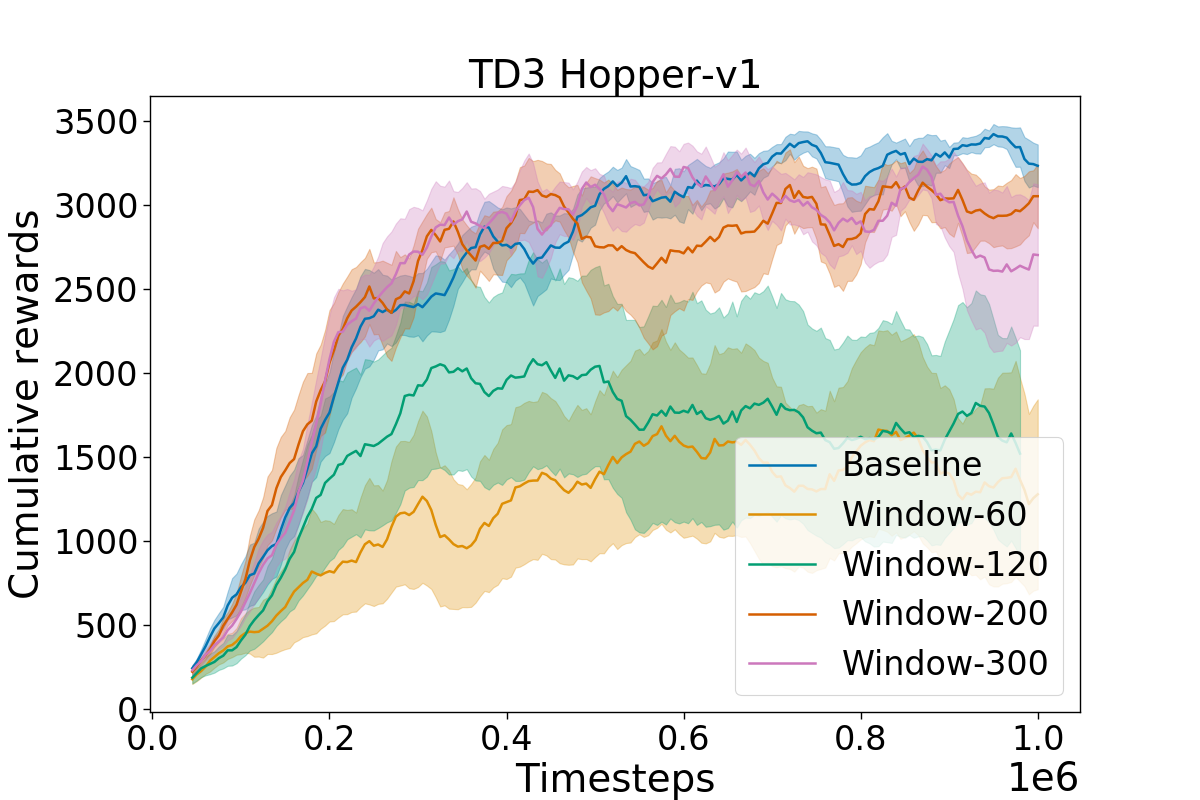}
    \includegraphics[width=.31\textwidth]{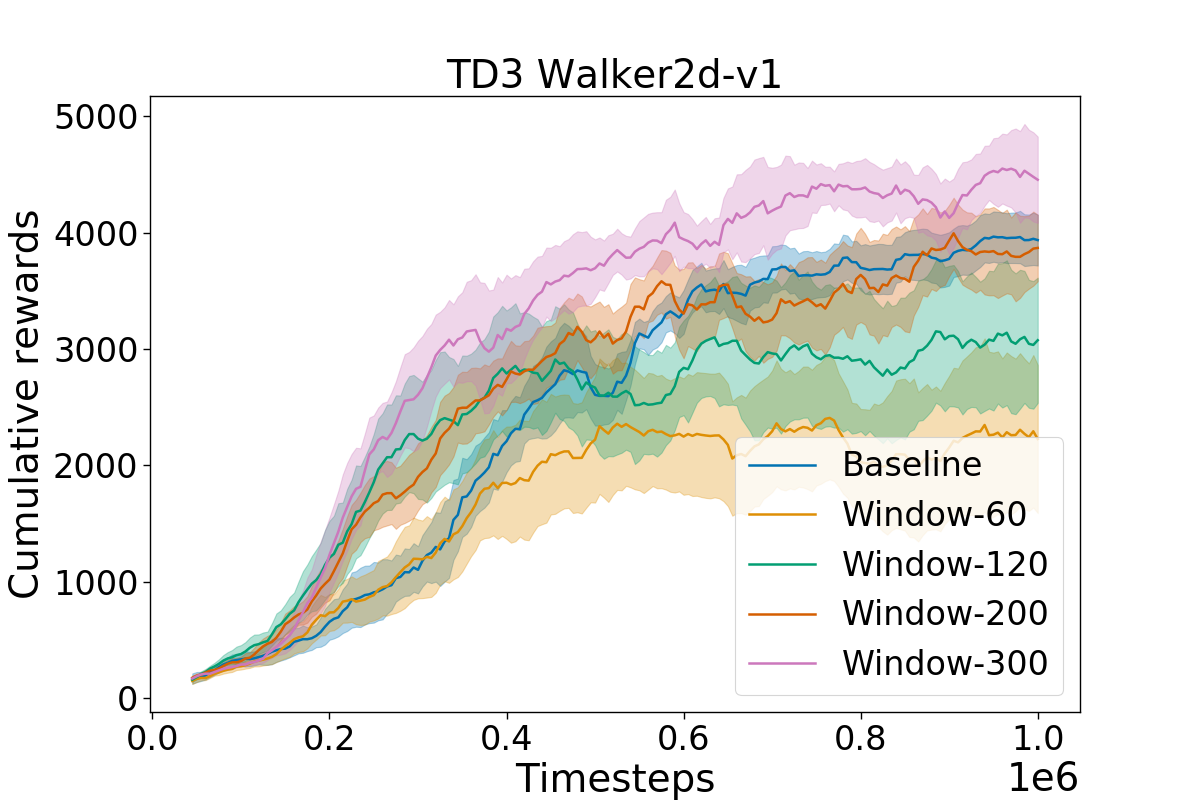}
    \caption{Reward curves for TD3 with windowed buffer sampling averaged over \textbf{10} seeds. Windowing is in episodes. The baseline corresponds to window size of $W = 10000$ episodes.}
    \label{fig:td3_window}
\end{figure*}

\textbf{Windowed Buffer : } Additional results for the \textit{windowed buffer} experiments with TD3 and SAC are shown in Figure \ref{fig:sac_window} and \ref{fig:td3_window}. Similar patterns as demonstrated for DDPG in the main draft hold for both these algorithms across all tasks. In particular, we see that environments with shorter buffer lengths (Hopper and Walker2d) perform worse with shorter windows. This is likely because they have sparsely distributed samples in the buffer, causing erroneous action value estimates.

\textbf{Summary - Windowed Buffer Experiments : } All our experiments show that with the buffer restricted to only the most recent samples, performance of the algorithms under study deteriorates. This suggests that they need data sufficiently spread across the state space for better generalization of action values.

\subsection{Ablation Studies with State Distribution Shift}

\begin{figure*}[!htb]
    \centering
    \includegraphics[width=.31\textwidth]{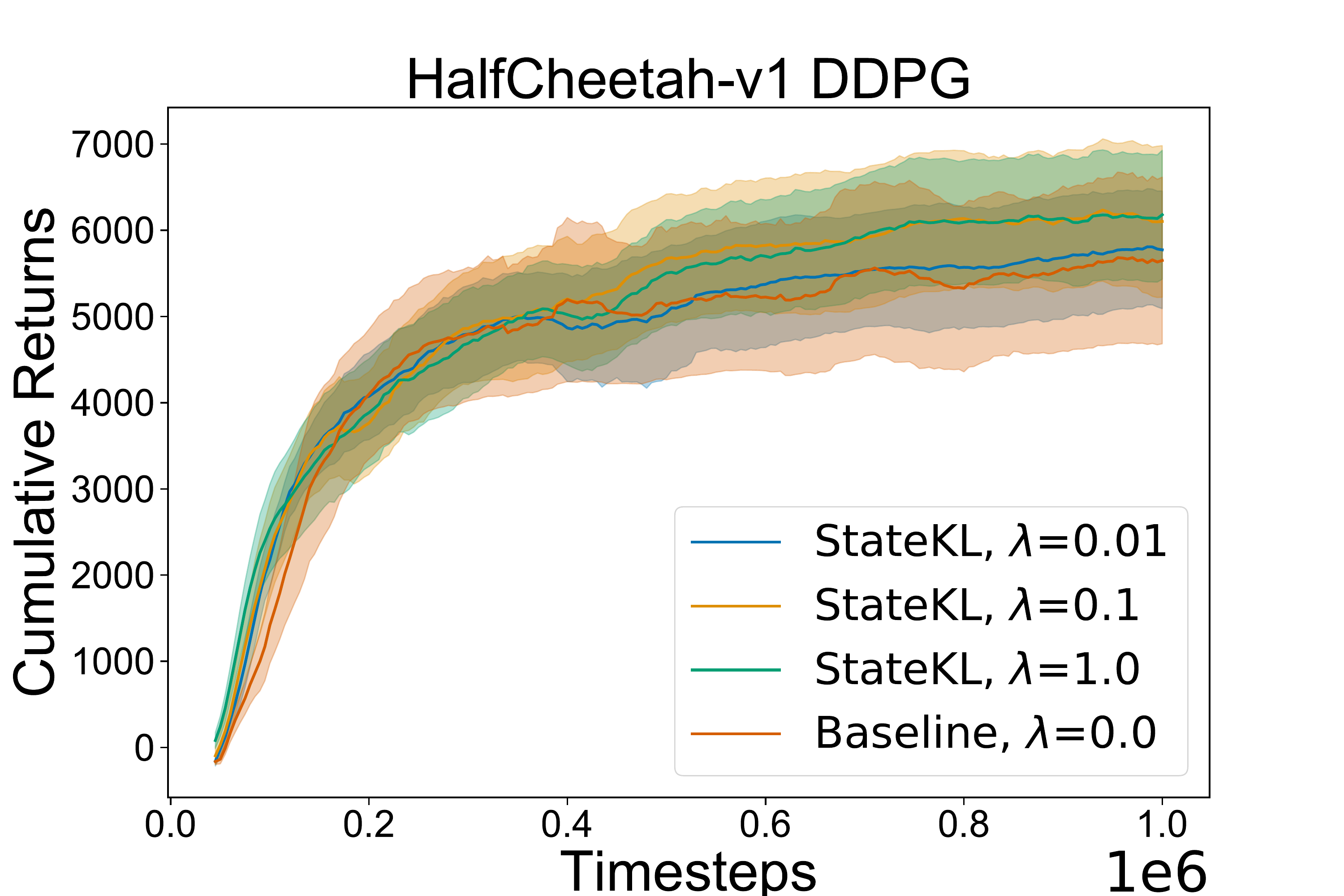}
    \includegraphics[width=.31\textwidth]{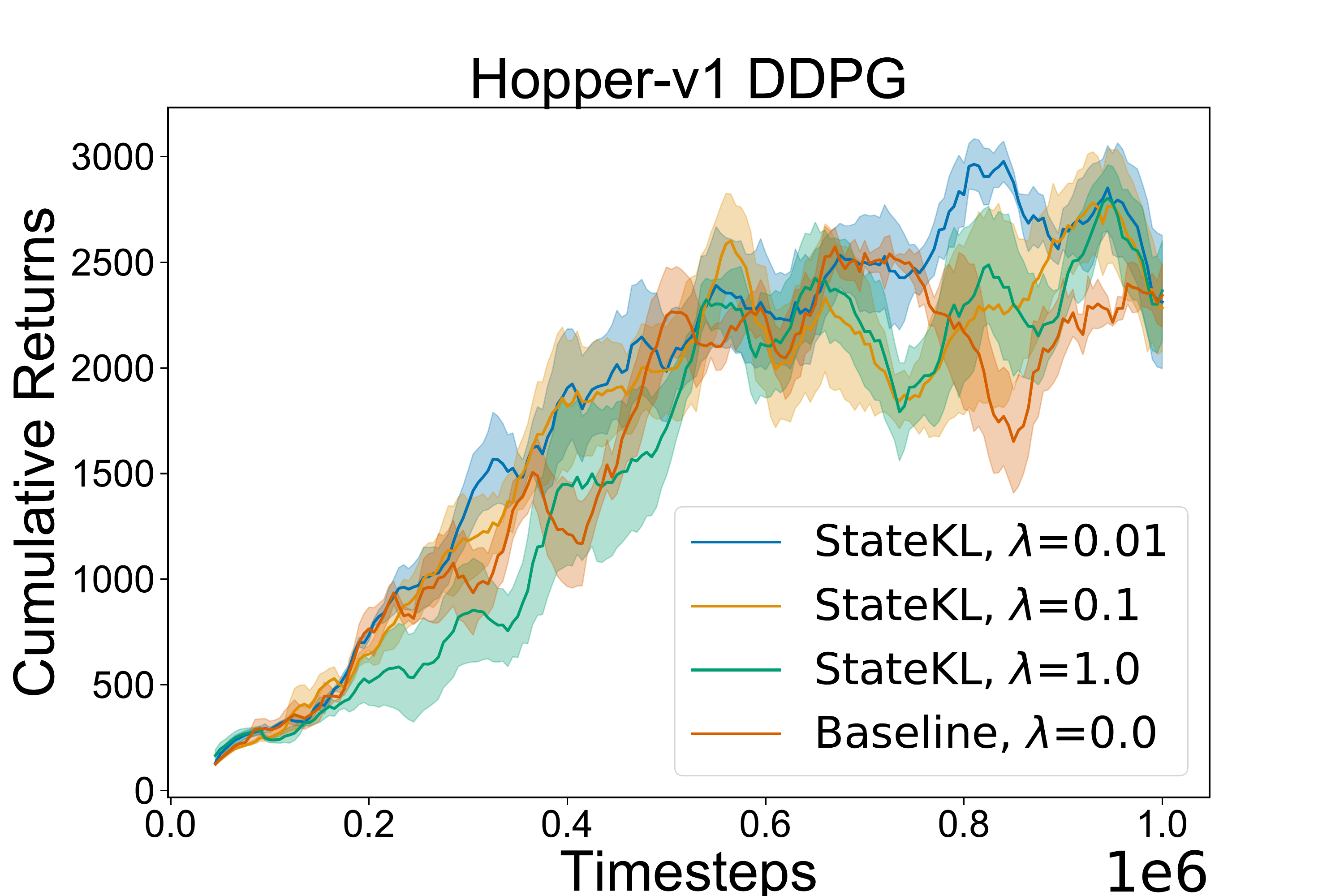}
    \includegraphics[width=.31\textwidth]{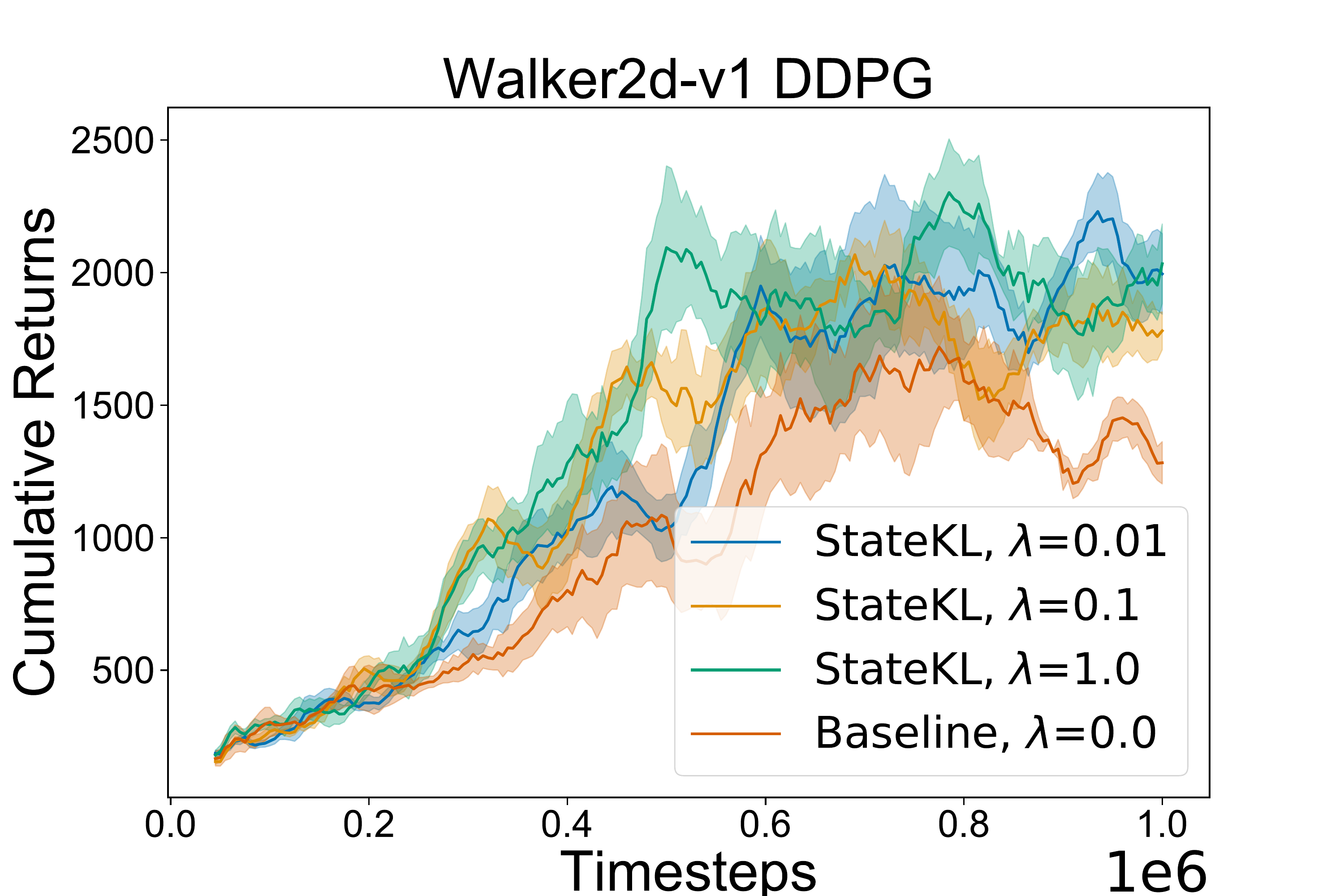}
    \caption{Ablation studies with different $\lambda$ StateKL regularizers in DDPG}
    \label{fig:DDPG_stateKL_ablation}
\end{figure*}

\begin{figure*}[!htb]
    \centering
    \includegraphics[width=.31\textwidth]{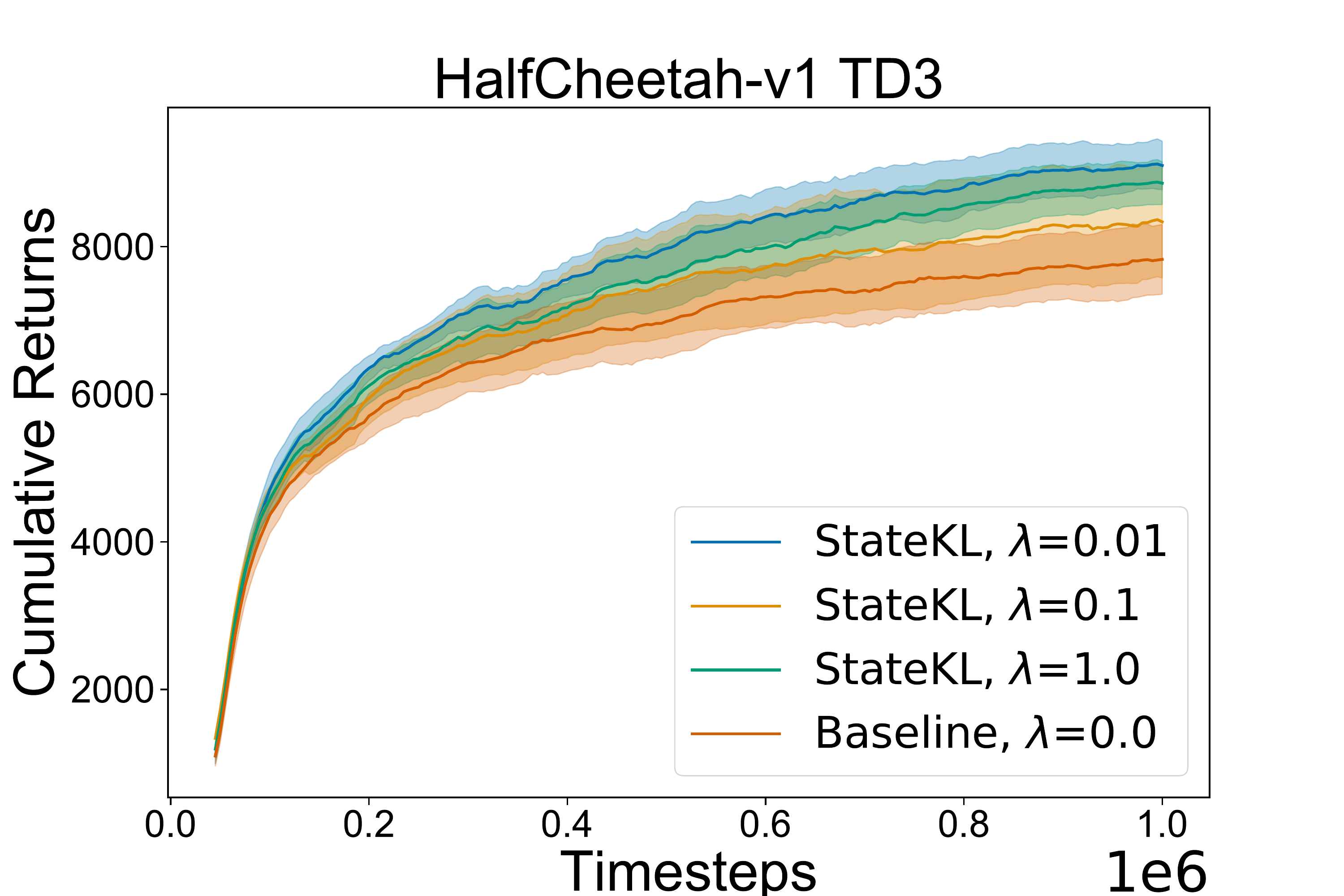}
    \includegraphics[width=.31\textwidth]{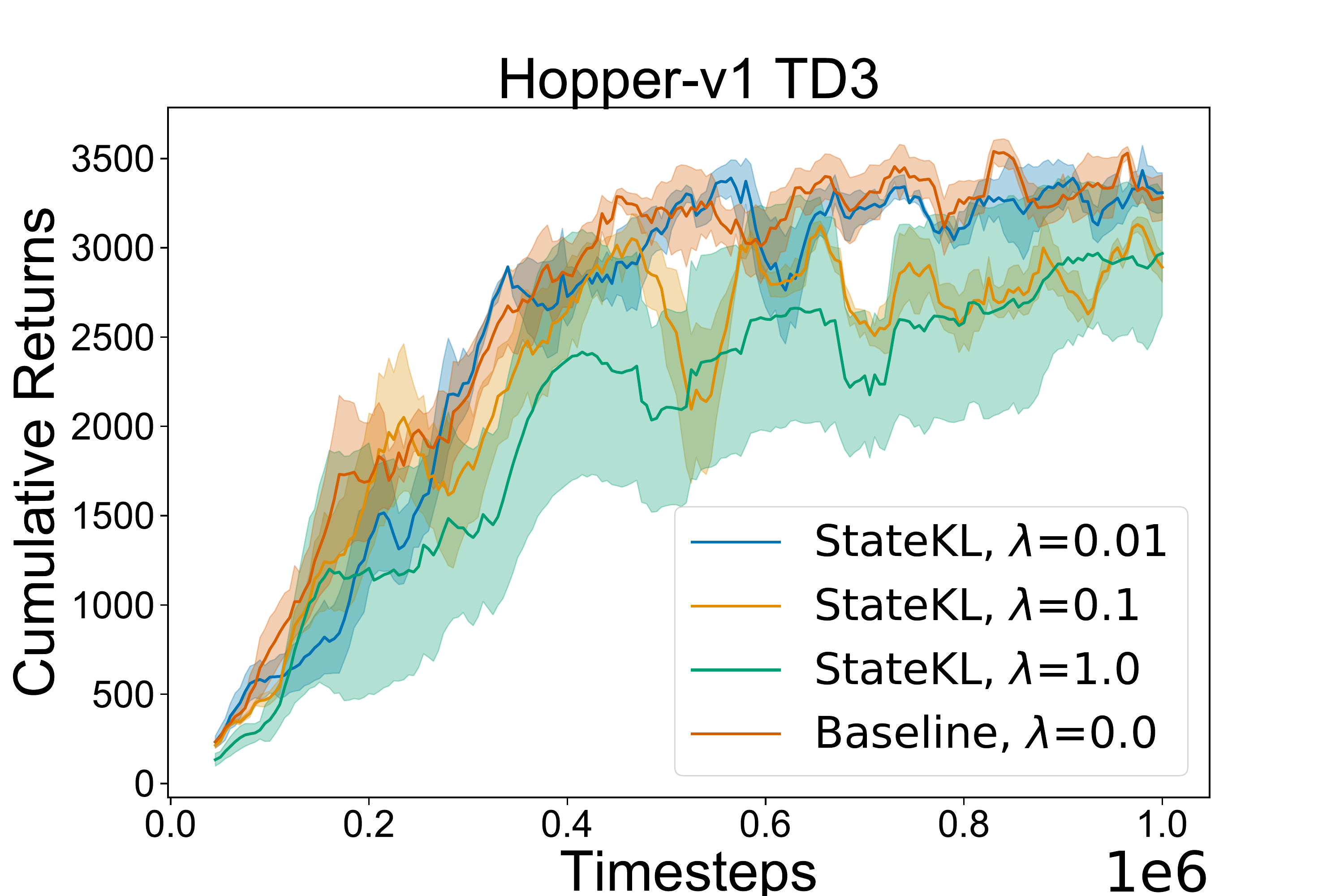}
    \includegraphics[width=.31\textwidth]{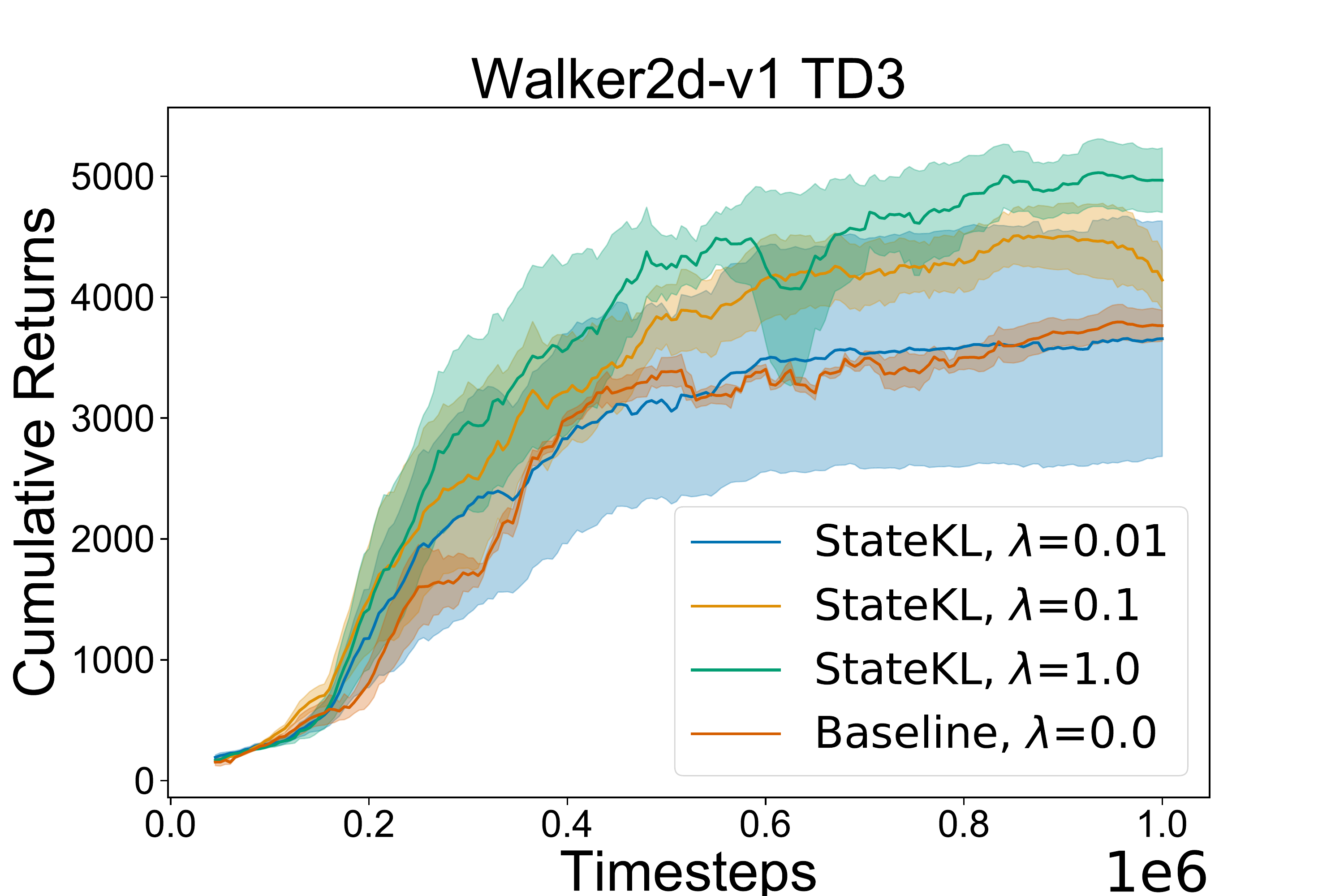}
    \caption{Ablation studies with different $\lambda$ StateKL regularizers in TD3}
    \label{fig:TD3_stateKL_ablation}
\end{figure*}

\begin{figure*}[!htb]
    \centering
    \includegraphics[width=.31\textwidth]{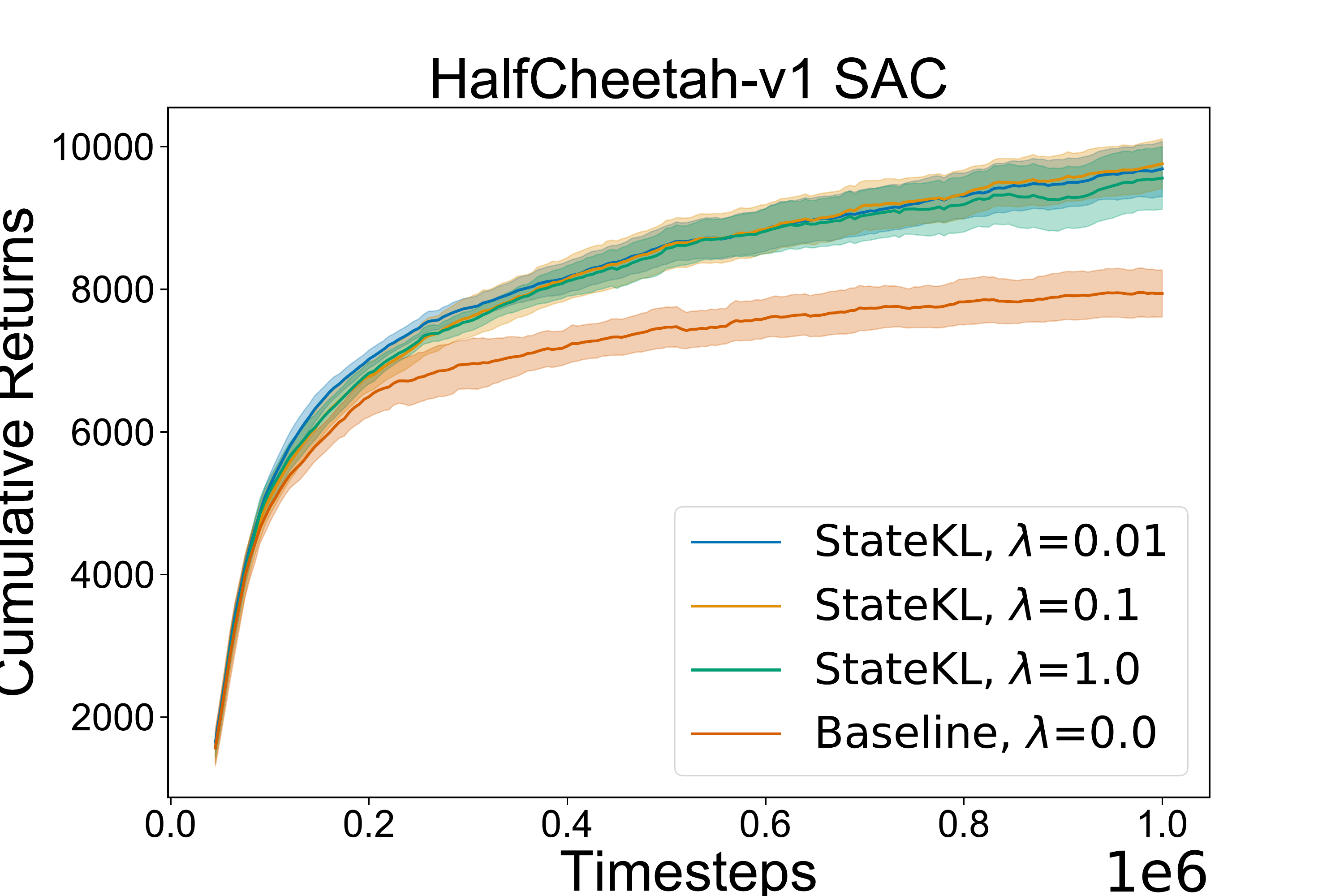}
    \includegraphics[width=.31\textwidth]{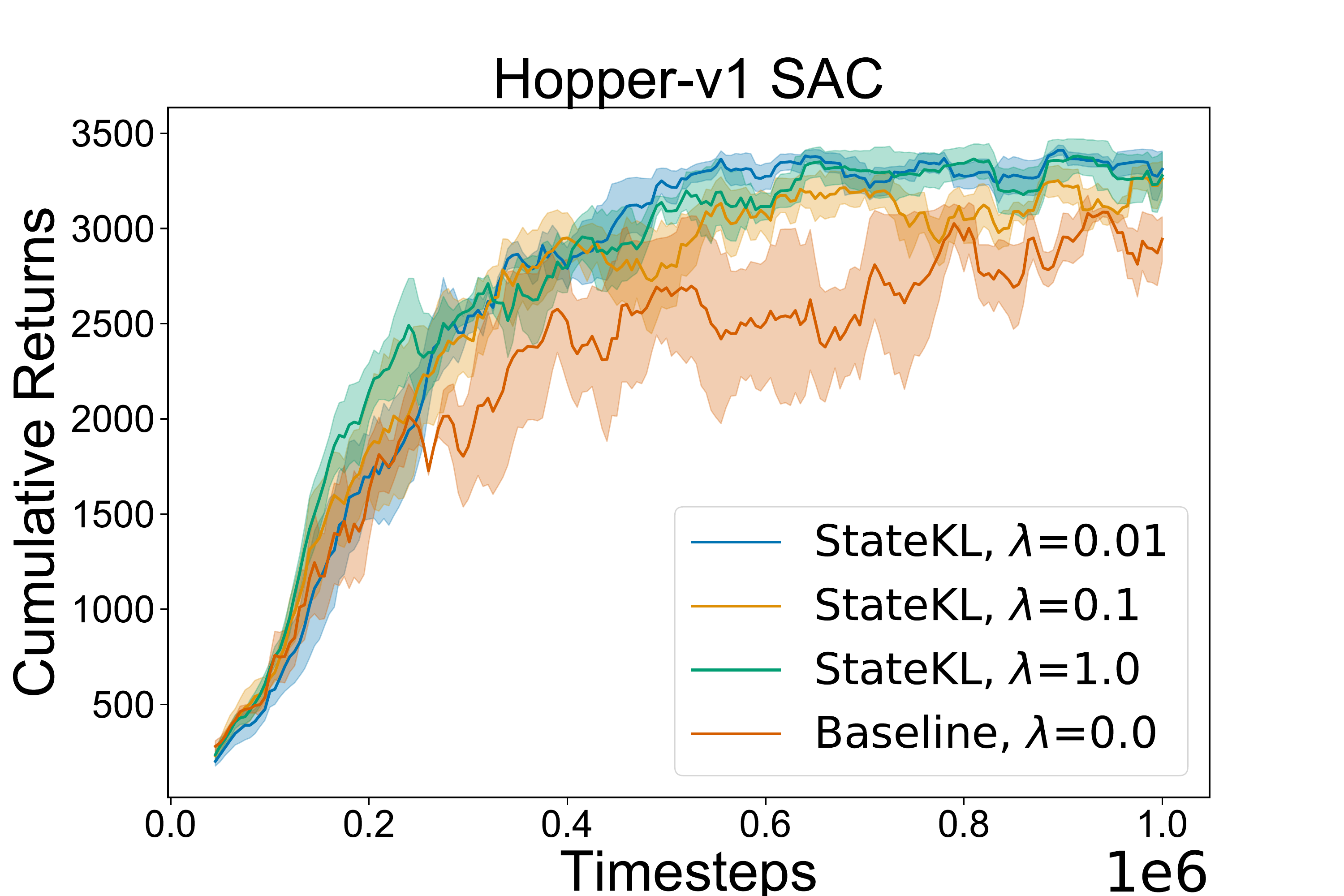}
    \includegraphics[width=.31\textwidth]{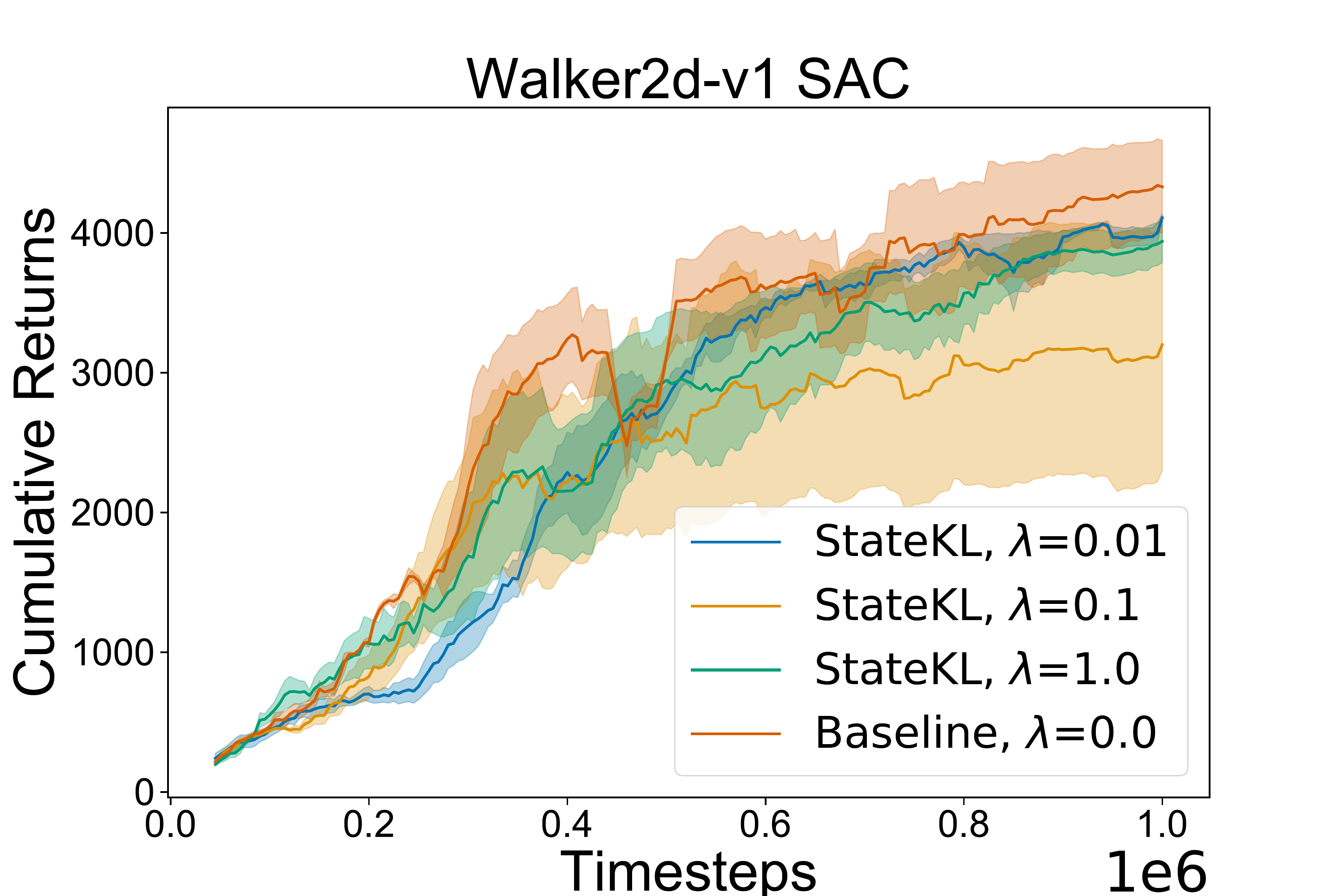}
    \caption{Ablation studies with different $\lambda$ StateKL regularizers in SAC}
    \label{fig:SAC_stateKL_ablation}
\end{figure*}

In this section, we include additional experimental results with different $\lambda$ weightings for the StateKL regularizer. We include results for a range of $\lambda$ and compare our results with baseline off-policy algorithms including DDPG, TD3 and SAC in Figures \ref{fig:DDPG_stateKL_ablation}, \ref{fig:TD3_stateKL_ablation}, and \ref{fig:SAC_stateKL_ablation}.


\end{document}